\documentclass[sigconf]{acmart}
\AtBeginDocument{%
  \providecommand\BibTeX{{%
    \normalfont B\kern-0.5em{\scshape i\kern-0.25em b}\kern-0.8em\TeX}}}

\usepackage{subfigure}
\usepackage{subcaption}
\usepackage{multirow}
\usepackage{float} 
\usepackage{amsmath}
\usepackage[ruled,linesnumbered]{algorithm2e}
\usepackage{graphicx}
\usepackage{subfigure}
\usepackage{epstopdf}
\usepackage{tablefootnote}
\usepackage[flushleft]{threeparttable}
\usepackage{diagbox}
\usepackage{booktabs}
\usepackage{siunitx}
\usepackage{footmisc}
\usepackage{footnote}
\usepackage{enumitem}
\usepackage[normalem]{ulem}
\useunder{\uline}{\ul}{}

\begin{document}

\title{CalibraEval: Calibrating Prediction Distribution to Mitigate Selection Bias in LLMs-as-Judges}

% \author{Anonymous Author(s)}
\author{Haitao Li}
\affiliation{DCST, Tsinghua University}
\affiliation{Quan Cheng Laboratory}
\email{liht22@mails.tsinghua.edu.cn}

\author{Junjie Chen}
\affiliation{DCST, Tsinghua University}
\affiliation{Quan Cheng Laboratory}
\email{chenjj24@mails.tsinghua.edu.cn}

\author{Qingyao Ai}
\affiliation{DCST, Tsinghua University}
\authornote{Corresponding author}
\affiliation{Quan Cheng Laboratory}

\email{aiqy@tsinghua.edu.cn}

\author{Zhumin Chu}
\affiliation{DCST, Tsinghua University}
\affiliation{Quan Cheng Laboratory}
% \affiliation{Beijing 100084, China}
\email{chuzm19@mails.tsinghua.edu.cn}

\author{Yujia Zhou}
\affiliation{DCST, Tsinghua University}
\affiliation{Quan Cheng Laboratory}
% \affiliation{Beijing 100084, China}
\email{zhouyujia@tsinghua.edu.cn}

\author{Qian Dong}
\affiliation{DCST, Tsinghua University}
\affiliation{Quan Cheng Laboratory}
% \affiliation{Beijing 100084, China}
\email{dq22@mails.tsinghua.edu.cn}

\author{Yiqun Liu}

\affiliation{DCST, Tsinghua University}
\affiliation{Zhongguancun Laboratory}
% \affiliation{Beijing 100084, China}
\email{yiqunliu@tsinghua.edu.cn}

\begin{abstract}
The use of large language models (LLMs) as automated evaluation tools to assess the quality of generated natural language, known as ``LLMs-as-Judges'', has demonstrated promising capabilities and is rapidly gaining widespread attention.
However, when applied to pairwise comparisons of candidate responses, LLM-based evaluators often exhibit selection bias. Specifically, their judgments may become inconsistent when the option positions or ID tokens are swapped, compromising the effectiveness and fairness of the evaluation result.
To address this challenge, we introduce CalibraEval, a novel label-free method for mitigating selection bias during inference. Specifically, CalibraEval reformulates debiasing as an optimization task aimed at adjusting observed prediction distributions to align with unbiased prediction distributions. 
To solve this optimization problem, we propose a non-parametric order-preserving algorithm (NOA).
This algorithm leverages the partial order relationships between model prediction distributions, thereby eliminating the need for explicit labels and precise mathematical function modeling.
Empirical evaluations of LLMs in multiple representative benchmarks demonstrate that CalibraEval effectively mitigates selection bias and improves performance compared to existing debiasing methods. This work marks a step toward building more robust and unbiased automated evaluation frameworks, paving the way for improved reliability in AI-driven assessments\footnote{The code can be found at \url{https://github.com/CSHaitao/CalibraEval}.}.

% This work will pave the way for more robust automated evaluation frameworks. It marks a significant step toward building more robust and unbiased automated evaluation frameworkss,
\end{abstract}
\keywords{LLM-as-Judges, Seclection Bias, Calibrate Prediction Distribution}

\maketitle

\section{Introduction}
In recent years, large language models (LLMs) have attracted widespread attention in both academia and industry~\cite{openai2023gpt4,zeng2022glm,blade}. These models achieve significant performance in a wide range of tasks, sometimes even exceeding human capabilities~\cite{RLCF}.
However, evaluating the quality of the texts generated by LLMs is difficult, particularly in subjective tasks such as open-ended story creation and summarization.
Traditional n-gram metrics (like BLEU~\cite{papineni2002bleu} and ROUGE~\cite{lin2004rouge}) and semantic-based metrics (such as BERTScore~\cite{zhang2019bertscore} and BARTScore~\cite{yuan2021bartscore}) are insufficient to comprehensively reflect the capabilities of LLMs.
Human evaluation, often regarded as the ``gold standard'', can measure model performance most accurately and provide valuable feedback, but it is costly and time-consuming.
Therefore, the demand for effective automated evaluation methods is growing increasingly~\cite{shi2024judging}.

\begin{figure}[t]
\vspace{-3mm}
\includegraphics[width=0.9\linewidth]{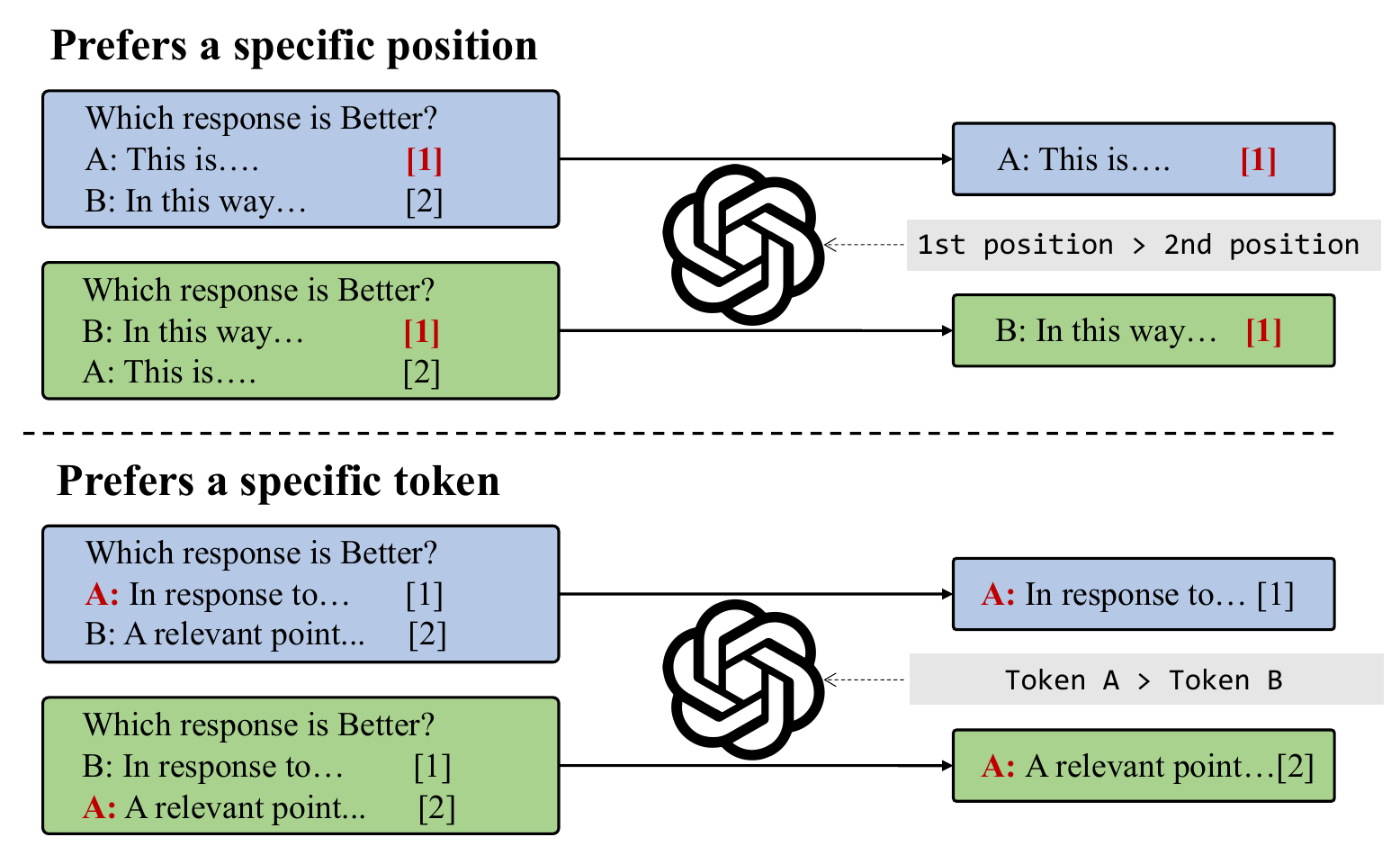}
\vspace{-3mm}
\caption{Illustration of selection bias in LLMs-as-Judges. Selection bias manifests in two aspects: prefers a specific position or prefers a specific token.}
\vspace{-6mm}
\label{bias}
\end{figure}

Some powerful commercial LLMs, such as GPT-4, have been widely applied to evaluate the quality of texts generated in response to open-ended questions. This paradigm, known as ``LLMs-as-Judges'', provides a scalable and transparent alternative to human evaluation of text quality. 
Within this paradigm, two common methods are pointwise and pairwise evaluations. In pointwise evaluation, LLMs assign scores to individual responses based on specific criteria, while in pairwise comparison, LLMs select the better response between two options.
Pointwise evaluation tends to be unstable and susceptible to noise, as subtle differences in wording or interpretation may lead to inconsistent results. In contrast, pairwise comparison can better reflect human judgment~\cite{liu2024aligning,zheng2023judging}, resulting in its widespread application and considerable attention.

Despite the success, LLMs are not perfect evaluators and are believed to exhibit certain biases~\cite{zheng2023large,zheng2023judging}. As shown in Figure ~\ref{bias}, when applied to pairwise comparisons of candidate responses, simply changing the positions or the ID tokens may lead to inconsistent evaluation results.
Previous studies have classified these biases as position bias~\cite{zheng2023judging,shi2024judging} and token bias~\cite{pezeshkpour2023large,raina2024llm}.
Positional bias refers to the tendency of LLMs to favor answers based on their specific positions (e.g., first or last), and token bias indicates that LLMs may assign more probability to certain option ID tokens (e.g., A or B).  
Given the inherent link between option tokens and their positions, we collectively refer to them as \textit{selection bias} in this paper.

Addressing selection bias in ``LLMs-as-Judges'' is crucial for ensuring valid and fair evaluations.
However, this task is not trivial, as selection bias is influenced by task-specific characteristics, such as domain and difficulty, as well as the inherent properties of LLMs, such as context window, family characteristics, and model capabilities~\cite{shi2024judging,ye2024justice}.
A straightforward method is to exclude inconsistent judgments or consider them ``ties''~\cite{chen2024humans,zheng2023judging}. While this approach enhances consistency and reliability, it may lead to a loss of evaluative information. Furthermore, more advanced methods, such as split and merge~\cite{li2023split} or discussions~\cite{chan2023chateval,li2023prd} among multiple agents, have been proposed to improve evaluation effectiveness.
However, these approaches typically require multiple rounds of interaction, making them costly and time-consuming, and their effectiveness in mitigating selection bias remains uncertain.

To address these limitations, we propose CalibraEval, a label-free, inference-time method for mitigating selection bias.
CalibraEval reformulates the debiasing problem as an optimization task to build a projection function that maps the original prediction distribution to an unbiased distribution. Our optimization objective is based on consistency judgments obtained after swapping option positions and ID tokens.
 Moreover, we propose a non-parametric order-preserving algorithm (NOA). The NOA narrows the solution space by preserving the partial order relationship between predicted distributions of observed samples. It derives the optimal calibration function by exploiting the relationship between the prediction distributions from different combinations of options. This approach effectively minimizes the reliance on explicit labels and precise mathematical function modeling, enhancing scalability and transferability.

We conduct extensive experiments on representative evaluation benchmarks with various LLMs. The experimental results indicate that CalibraEval outperforms strong baselines in debiasing performance and achieves state-of-the-art results. Furthermore, we validate CalibraEval's robustness across diverse prompt templates, varied option tokens, and in-context learning scenarios, demonstrating its potential for application in a variety of contexts.
To summarize, we make the following contributions: 

\begin{enumerate} 
\item We propose a label-free, inference-time calibrated method CalibraEval. By learning a lightweight calibration function, CalibraEval effectively mitigates selection bias, demonstrating both significant effectiveness and efficiency.
\item We reformulate the debiasing problem as an optimization task and propose a non-parametric order-preserving algorithm (NOA) to solve it efficiently.
% reduce the need for explicit label and precise mathematical function modeling.
\item We conduct extensive experiments on public benchmarks. Experimental results demonstrate the effectiveness and robustness of CalibraEval.
\end{enumerate}

\section{Related Work}

\subsection{LLMs as Judges}
The rapid development of large language models (LLMs) in recent years has highlighted the urgent need for effective evaluation methods~\cite{shi2024judging,wei2024unveiling,Sailer,chen2024automaticcostefficientpeerreviewframework}. 
Traditional evaluation metrics, such as BLEU~\cite{papineni2002bleu} and ROUGE~\cite{lin2004rouge}, fall short in comprehensively capturing model performance. 
These metrics typically overlook nuanced aspects of generated texts, such as coherence, relevance, and contextual appropriateness.
Moreover, manual evaluation can provide more accurate assessments and nuanced insights, but it is both costly and time-consuming, making it impractical for large-scale assessments. 
This situation highlights the urgent need for more advanced and efficient automated evaluation techniques that can keep pace with the evolving capabilities of LLMs~\cite{chu2024prepeerreviewbased,T2ranking,Lecardv2,li2024lexevalcomprehensivechineselegal}.

To tackle these challenges, the ``LLMs-as-Judges'' approach has emerged as a promising alternative~\cite{ye2024justice}. This method utilizes powerful, widely recognized LLMs, such as GPT-4~\cite{openai2023gpt4}, to facilitate automated evaluation, thereby reducing the dependence on manual assessment.
Generally speaking, the ``LLMs-as-Judges'' evaluation approach can be classified into two categories: pointwise~\cite{kim2023prometheus} and pairwise~\cite{lambert2024rewardbench,zhu2023judgelm}. Pointwise evaluation involves LLM judges scoring individual responses based on specific criteria. Pairwise comparison requires choosing the better answer from two responses.
Pairwise comparison evaluation has gained widespread adoption and particular attention due to its outstanding performance.
Wang et al. ~\cite{wang2023largelanguagemodelsfair} discovered that pairwise comparison methods outperform traditional score-based evaluation approaches in terms of consistency with human assessments. Liu et al.~\cite{liu2024aligning} observed that pairwise comparisons better reflect human evaluation standards compared to other methods. This advantage may be attributed to the fact that LLMs often utilize pairwise preference or ranking data during the Reinforcement Learning from Human Feedback (RLHF) training phase~\cite{dong2024unsupervisedlargelanguagemodel}.

Furthermore, some researchers have explored integrating multiple LLMs into evaluation systems, aiming to produce effective results through collaboration~\cite{chan2023chateval}, discussion~\cite{li2023prd}, and debate~\cite{chu2024prepeerreviewbased} among the models. However, these approaches typically require multiple rounds of interaction, leading to increased resource consumption. 
In summary, pairwise comparison evaluation is relatively more straightforward and less resource-intensive, which is regarded as a more economical and effective solution.

\subsection{Bias in LLM Judges}
While ``LLMs-as-Judges'' has emerged as a promising alternative to human evaluation in many tasks, concerns have been raised about the reliability of these judges due to potential biases inherent in LLMs. 
These biases pose significant challenges to the effectiveness and fairness of such evaluation systems~\cite{shi2024judging,zhou2024mitigating}.

Recent research has identified various biases affecting LLM evaluations, including selection bias~\cite{zheng2023judging}, position bias~\cite{li2023split}, contextual bias~\cite{zhou2023batch}, and self-reinforcing bias~\cite{li2023prd,zhang2024evaluationethicsllmslegal}. Among these, selection bias has emerged as a particularly critical issue, as it is prevalent across various tasks and affects both open-source and commercial models, significantly impacting their performance. This bias is typically evident in pairwise comparison evaluations: if an LLM evaluation model consistently favors a specific option even after swapping positions or IDs, this indicates the presence of selection bias. Two types of bias may contribute to selection bias: position bias and token bias. However, there is still no consensus on which of these two is the dominant factor~\cite{pezeshkpour2023large,raina2024llm}.

% To address these biases, researchers have proposed and implemented several mitigation strategies.
Effectively mitigating bias remains an unresolved issue. Scholars are exploring various approaches to identify and reduce biases in LLMs~\cite{shi2024judging,zhou2024mitigating,zhang2024evaluationethicsllmslegal}.
Shi et al.~\cite{shi2024judging} conducted a systematic study on positional bias through pairwise comparison evaluations, providing detailed recommendations for selecting judgment LLMs that balance consistency, fairness, and cost-effectiveness. Chua et al. ~\cite{chua2024bias} proposed Bias-Consistent Training (BCT) to fine-tune models, aiming to enhance consistent reasoning between prompts with or without biased features. Li et al.~\cite{li2023split} introduced the ``split and merge'' method, which divides answers into multiple parts and aligns similar content in candidate answers to calibrate position bias. Zheng et al. ~\cite{zheng2023large} use prior estimates from partial samples to address selection bias. Furthermore, Liu et al. ~\cite{liu2024aligning} argued that existing calibration techniques aimed at reducing bias are insufficient for calibrating LLM evaluators, even with supervised data.
Therefore, mitigating biases in ``LLM-as-Judges'' is a widespread, significantly impactful, and challenging issue.

\section{CalibraEval}

In this section, we first present the problem statement of the debiasing process and the optimization objective. Then, we provide a detailed introduction to the non-parametric order-preserving algorithm (NOA).

\subsection{Problem Statement }

In this paper, we focus on addressing the selection bias present in ``LLMs-as-Judges''. Selection bias refers to the phenomenon where LLMs consistently prefer a specific option during pairwise comparisons, regardless of the content.

To standardize the terminology, we define the following terms: 
$t_i$ represent the option ID tokens (e.g., A, B), and $o_i$ denotes the specific option contents (e.g., Response\_x, Response\_y). Additionally, let $I$ represent the input instruction, and $X_0$ represent the default connection of option ID tokens and contents, that is, $X_0=[(t_1,o_1);(t_2,o_2)]$.

Following previous studies~\cite{zheng2023large}, we assume that when the LLM serves as an evaluator, the observed probability distribution $P_{observed}$ on $t_i$ can be decomposed into a combination of the prior distribution $P_{prior}$ and the debiased distribution $P_{debiased}$, i.e.,
\begin{equation}\label{eqn-1} 
P_{observed}(t_i|I,X_0) = f(P_{prior}(t_i|I,X_0),P_{debiased}(t_i|I,X_0) ) 
\end{equation}
where $f(\cdot)$ is a function that represents the relationship between $P_{observed}$, $P_{prior}$ and $P_{debiased}$.
Accurately estimating the form of $f(\cdot)$ is challenging. Firstly, the interaction between $P_{prior}$ and $P_{debiased}$ is complex and may not be simply multiplicative or additive.
Secondly, the observed probability distributions $P_{observed}$ may be affected by noise, complicating the identification of the precise form of $f(\cdot)$.
In previous work, Zheng et al.~\cite{zheng2023large} proposed Pride, which simplify the problem by assuming that $f(\cdot)$ is a linear multiplication, i.e.,
\begin{equation}\label{eqn-2} 
P_{observed}(t_i|I,X_0) \propto Z^{-1}_{I,X_0} P_{prior}(t_i|I,X_0) \times P_{debiased}(t_i|I,X_0)
\end{equation}
where $Z^{-1}_{I,X_0}$ is the normalization item. Zheng et al.~\cite{zheng2023large} select a subset of test samples and then use the average observed probability distributions from different arrangements as the prior estimates $\tilde{P}_{prior}(t_i)$.
The debiasing is then performed using the following equation:
\begin{equation}\label{eqn-3} 
 P_{debiased}(t_i|I,X_0) \propto P_{observed}(t_i|I,X_0) / \tilde{P}_{prior}(t_i)
\end{equation}

Although Pride is effective, its simplified assumption overlooks the complex relationships between probability distributions, leading to suboptimal performance.

In this paper, considering the complexity of $f(\cdot)$, 
we do not attempt to directly create a precise mathematical function of $f(\cdot)$.
Instead, we focus on determining a calibration function $g(\cdot)$, which can map the observed probabilities to an unbiased probability distribution, i.e.,
\begin{equation}\label{eqn-4} 
 P_{debiased}(t_i|I,X_0) = g(P_{observed}(t_i|I,X_0))
\end{equation}

\begin{figure}[t]
\vspace{-3mm}
\includegraphics[width=\linewidth]{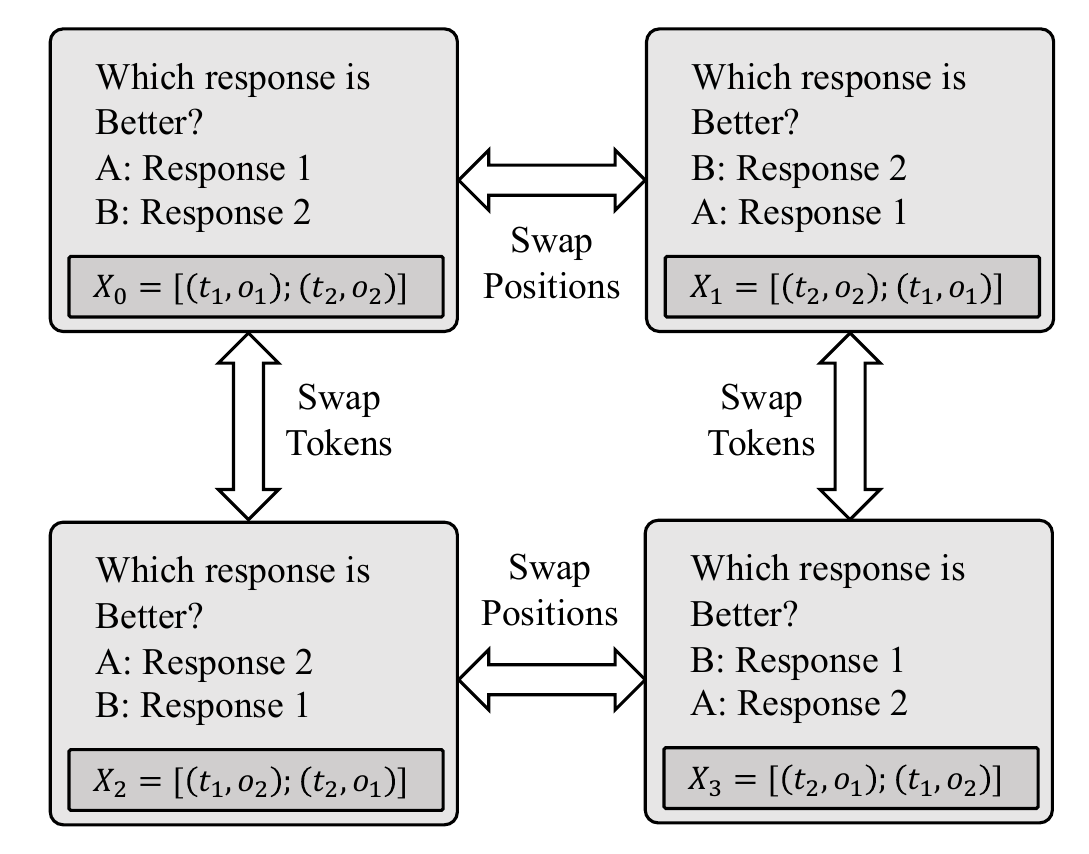}
\vspace{-5mm}
\caption{Four different types of combinations. $t_1/t_2$ represents the option IDs (A/B), while $o_1/o_2$ denotes the corresponding option contents.
An unbiased evaluator consistently ranks the responses regardless of changes in option order (Swap Positions) or option ID tokens (Swap Tokens), ensuring fairness and consistency in the results.}
\vspace{-5mm}
\label{judges}
\end{figure}

\subsection{Optimization Objective}
In this section, we reformulate the debiasing problem as an optimization task, with the unbiased probability distribution serving as the optimization objective.
Intuitively, an unbiased evaluator should provide consistent judgments even when the option position or ID tokens are swapped.
Specifically, in pairwise comparisons, there are four possible combinations of positions and ID tokens: 
\begin{equation}\label{eqn-5} 
X_0=[(t_1,o_1);(t_2,o_2)],  X_1=[(t_2,o_2);(t_1,o_1)]
\end{equation}
\begin{equation}\label{eqn-6} 
X_2=[(t_1,o_2);(t_2,o_1)],  X_3=[(t_2,o_1);(t_1,o_2)]
\end{equation}

In Figure~\ref{judges}, we present the relationship among these four combinations. An unbiased evaluator can accurately determine the correct option context, regardless of changes in option orders (Swap Positions) or option ID tokens (Swap Tokens). Suppose that the ground truth is $o_1$, the evaluator should satisfy the following conditions:
\begin{equation}\label{eqn-7} 
P_{debiased}(t_1|I,X_0) = P_{debiased}(t_1|I,X_1)=P_{debiased}(t_2|I,X_2)
\end{equation}
\begin{equation}\label{eqn-8} 
P_{debiased}(t_2|I,X_2) = P_{debiased}(t_2|I,X_3)=P_{debiased}(t_1|I,X_0)
\end{equation}

Since Equations (\ref{eqn-7}) and Equations (\ref{eqn-8}) are duals, 
we only need to select one as the optimization objective.
Also, we can simply normalize the original token prediction probabilities, ensuring that the sum of the probabilities for outputs $t_1$ and $t_2$ equals 100\%, i.e.,
\begin{equation}\label{eqn-9} 
P_{debiased}(t_1|I,X_0) = 1 - P_{debiased}(t_2|I,X_0)
\end{equation}

With the above reasoning, we formulate the debiasing problem on $K$ samples as follows:
\begin{equation}\label{eqn-10} 
\min_{g\in \mathcal{G}}  \sum_{i=1}^{K}[ g(s_{0}^{i}) + g(s_2^i)-1]^2 +[ g(s_{0}^{i}) - g(s_1^i)]^2-\lambda[ g(s_{0}^{i}) - g(s_2^i)]^2
\end{equation}
\begin{equation}\label{eqn-11} 
s.t. s_{j}^{i} = P_{observed}(t_1|I,X_j), j={0,1,2}, i={1,...,K}.
\end{equation}
where $g(\cdot)$ is the mapping function for the probability of token $t_1$. $\mathcal{G}$ denotes the solution space of $g(\cdot)$.
$\lambda$ is a hyper-parameter.
For each option ID token,  a corresponding mapping function $g(\cdot)$ is defined. In the following process, we use $g(\cdot)$ as an example.
In Equations (\ref{eqn-10}), the first term ensures consistent judgments when option ID tokens are swapped. The second term aims to maintain consistent judgment when option positions are exchanged. The third term serves as a regularization term, which prevents convergence to the trivial solution $g(\cdot) = 0.5$.

\subsection{Non-parametric Order-Preserving  Algorithm (NOA)}
The optimization problem presented in Equation (\ref{eqn-10}) is an NP problem, featuring an extensive solution space $\mathcal{G}$. Furthermore, the absence of explicit labels prevents us from employing supervised methods to determine $g(\cdot)$.

To address these limitations, we propose a non-parametric order-preserving algorithm called NOA. Non-parametric methods do not rely on specific model assumptions, making them well-suited for handling high-dimensional data or complex functions.  NOA searches for the optimal solution by directly evaluating the output of the calibration function, eliminating the need for explicit labels or precise mathematical modeling.

To narrow the solution space $\mathcal{G}$, we assume that the mapping function $g(\cdot)$ is order-preserving for the same ID token. 
This assumption, widely and implicitly applied in previous work~\cite{zheng2023large,zhao2021calibrate}, rests on the premise that the prior distribution $P_{prior}$ reflects the LLM's inherent bias toward certain option ID tokens, which remains conditionally independent of the unbiased probability distribution $P_{debiased}$.
Intuitively, for a given LLM, the partial order relationship under the same prior bias should remain consistent, meaning higher observed probabilities generally correspond to higher unbiased probabilities for the same ID token.

Specifically, we first collect an estimation set with $K$ samples.
Each sample is processed by swapping ID tokens and swapping positions, resulting in three probabilities $s_0$ (default output), $s_1$ (swap positions), and $s_2$ (swap ID tokens).
The probabilities from all samples are combined into a set $S = \{s^i_0, s^i_1, s^i_2 | i \in {1, ..., K}\}$. Then, we sort $S$ in ascending order to form a sequence $z_1 \leq z_2 \leq ... \leq z_{M-1}$, where $M = 3K + 1$. We then append boundary conditions to the sorted sequence by defining $z_0=0$ and $z_{M} = 1$, producing the complete sequence $Z = \{z_0,z_1,...,z_{M-1},z_{M}\}$.

To optimize the model, we introduce a set of parameters $d_k$ ($k=0,1,2, ..., M$) initialized to the values of $z_k$.
These parameters will be optimized during the process.
Then, we define the mapping function $g(\cdot)$ using the softmax-like expression:
\begin{equation}\label{eqn-12} 
g(z_k) = \frac{\sum_{i=0}^{k}exp(d_i) }{  {\textstyle \sum_{i=0}^{M}} exp(d_i)}
\end{equation}

$g(\cdot)$  is a discrete mapping function with parameters $d_k$, which satisfies the constraint of order preservation.
We employ gradient descent methods to iteratively update the parameters $d_k$. The update rule is given by:

\begin{equation}\label{eqn-13} 
d_k^{(new)} = d_k^{(old)} - \gamma \frac{\partial L}{\partial d_k} 
\end{equation}
where $\gamma$ is the learning rate, $L =[g(s_{0}^{i}) + g(s_2^i)-1]^2+[ g(s_{0}^{i}) - g(s_1^i)]^2-\lambda[ g(s_{0}^{i}) - g(s_2^i)]^2$. 
This iterative process allows the parameters to converge toward the optimal values that minimize the loss, thereby reducing the bias in the probability distribution.

For $\frac{\partial L}{\partial d_{k}}$, we derive the following equation. The detailed derivation process can be found in Appendix ~\ref{proof}.
\begin{align}\label{eqn-14} 
\frac{\partial L}{\partial d_{k}} & = \left(2\left[g\left(s_{0}^{i}\right)+g\left(s_{2}^{i}\right)-1\right]+2\left[g\left(s_{0}^{i}\right)-g\left(s_{1}^{i}\right)\right]\right) \frac{\partial g\left(s_{0}^{i}\right)}{\partial d_{k}} \nonumber \\
& \quad + \left(-2\left[g\left(s_{0}^{i}\right)-g\left(s_{1}^{i}\right)\right]\right) \frac{\partial g\left(s_{1}^{i}\right)}{\partial d_{k}}  \nonumber \\
& \quad + \left(2\left[g\left(s_{0}^{i}\right)+g\left(s_{2}^{i}\right)-1\right] - 2 \lambda\left[g\left(s_{0}^{i}\right) - g\left(s_{2}^{i}\right)\right]\right) \frac{\partial g\left(s_{2}^{i}\right)}{\partial d_{k}} 
\end{align}

\begin{equation} \label{eqn-15}
\frac{\partial g\left(z_{j}\right)}{\partial d_{k}} =
\begin{cases}
-\frac{\sum_{i=0}^{j} \exp \left(d_{i}\right) \exp \left(d_{k}\right)}{\left(\sum_{i=0}^{M} \exp \left(d_{i}\right)\right)^{2}} & (j < k) \\
\frac{\exp\left(d_{k}\right) \left( \sum_{i=0}^{M} \exp\left(d_{i}\right) - \sum_{i=0}^{j} \exp \left(d_{i}\right) \right)}{\left(\sum_{i=0}^{M} \exp\left(d_{i}\right)\right)^{2}} & (j \ge k)
\end{cases}
\end{equation}

We note that there are infinite solutions that satisfy the optimization problem.
This is because any constant change in the value of $d_k$ does not affect the relative values in the exponential terms of Equation (\ref{eqn-12}).
To obtain a unique solution, we apply the normalization constraint $\sum_{i=0}^{M}d_i = 0$ after each iteration.
The optimization proceeds until a convergence criterion is met, such as the loss function $L$ reaching a minimum threshold or the parameter updates becoming sufficiently small.

After the solution process converges, we obtain the sample points $Z = \{z_1,...,z_{M-1}\}$ and their corresponding calibrated values $y = \{g(z_1),...,g(z_{M-1}\}$. 
For sample points not included in $Z$, we use existing sample points to learn the continuous calibration function $g^*(\cdot)$.
The goal is to identify a set of non-decreasing piecewise linear functions that minimize the sum of squared deviations between the estimated values and the calibrated values of the samples.
Specifically, we fit the calibration values by minimizing the following objective function:

\begin{equation} \label{eqn-16}
min \sum_{i=1}^{M-1}w_i(g(z_i)-g^*(z_i))^2 
\end{equation}
\begin{equation} \label{eqn-17}
s.t. z_1 \le z_2 ... \le z_{M-1} 
\end{equation}
\begin{equation} \label{eqn-18}
\sum_{i=1}^{M-1}w_i = 1, w_i \ge 0
\end{equation} 
The above problem is a weighted least squares quadratic programming problem. We apply the Pool Adjacent Violators Algorithm (PAVA)~\cite{zadora2014pool} to derive the continuous calibration function $g^*(\cdot)$.

It is worth noting that CalibraEval does not require explicit labels and can be executed during inference with minimal computational cost. The calibration function can be calculated after observing all test samples or by utilizing a subset of samples.
The entire process of CalibraEval is summarized in Algorithm \ref{algo} in the Appendix.

% \begin{algorithm}[t]
% 	\caption{Calibration process of ClibraEval}
% 	\label{Retrival}
% 	\KwIn{Language model, test samples, 
% 	\KwOut{$k$ nearest approximate candidates }
    
%     Result set $A=\emptyset$, candidate set $Z$=\{the root node $n_1$\}.
    
%     \While{$Z \not =\emptyset$}{
%     Remove all leaf nodes from $Z$ and insert them into $A$.
%     % Move the leaf nodes in $Q$ to set $A$.
    
%     Calculate $s=\widetilde{e}_{c_n}^T \cdot \Phi(q)$ for each remaining node $n\in Z$.  ${e}_{c_n}$ is the embedding of current node.
    
%     According to the $s$, top $(b-len(A))$ nodes in $Z$ are selected to form the set $I$. There are no leaf nodes in $I$.
    
%     Update the $Z$: $Z$ = \{children nodes of $n|n \in I$\}.

%     }
%     Compute $s=\widetilde{e}_{d}^T \cdot \Phi(q)$ for documents $d$ contained by nodes in $A$ to get top $k$ candidates.
% \end{algorithm}

\section{EXPERIMENT SETUP}
% In this section, we first introduce our experimental setup, including datasets and metrics, baselines, and implementation details. Then, we presented the experimental results and conducted a detailed analysis.

\subsection{Datasets and Metrics}
We conduct experiments on three representative benchmarks. The statistics are shown in Appendix \ref{data_detail}.
\begin{itemize}[leftmargin=*]
    \item \textbf{RewardBench}~\cite{lambert2024rewardbench} is a benchmark dataset designed for evaluating reward models. It contains 2,985 prompt-choice-rejection trios across four task categories: Chat, Chat Hard, Safety, and Reasoning.
    \item \textbf{MTBench}~\cite{zheng2023judging} is a multi-turn response dataset. It contains 3,355 expert-level pairwise human preferences for responses, generated by 6 models for 80 MTBench questions.
    \item \textbf{PreferenceBench}~\cite{kim2024prometheus} is a test set designed to assess the evaluation capabilities of LLMs, comprising 2,000 response pairs (classified as ``win'' or ``lose'') and 200 evaluation criteria.
\end{itemize}

In the evaluation, we primarily utilize \textbf{reference-free} metrics to measure the consistency of model evaluations. We compute Fleiss's Kappa coefficient~\cite{falotico2015fleiss} and intraclass correlation coefficient (ICC)~\cite{bartko1966intraclass} between the evaluation results obtained after swapping option ID tokens and option positions. We report two specific ICC metrics: ICC(2,k) and ICC(3,k) in this paper.

For the \textbf{reference-based} evaluation, we report the standard deviation of recalls (RStd) and accuracy. Following Zheng et al.~\cite{zheng2023large}, the balance of recalls serves as an effective measure of the extent of selection bias. A greater imbalance in recalls signifies a more pronounced selection bias. In addition, MTBench includes ``tie'' options assessed by human evaluators. We exclude all ``tie'' options when calculating the reference-based metrics.
In Appendix \ref{metric_detail}, we provide the details of evaluation metrics.

% We employ two types of evaluation metrics: label-independent and label-irrelevant metrics. For label-independent metrics, we compute Fleiss' Kappa~\cite{falotico2015fleiss} coefficient and the Intraclass Correlation Coefficient (ICC)~\cite{bartko1966intraclass} across various context permutating to assess the consistency of evaluation outcomes. 
% We report on ICC(2,k) and ICC(3,k). ICC(2,k) measures the consistency of ratings from multiple raters for the same set of subjects under a random effects model, whereas ICC(3,k) assesses the consistency of ratings from specific and fixed raters for the same subjects under a fixed effects model.
% Regarding label-irrelevant metrics, we introduce the standard deviation of recalls (RStd) as a quantitative measure, where a larger recall imbalance indicates a more pronounced selection bias. Furthermore, we include Accuracy as one of the supplementary metrics in our reporting.

\begin{table*}[]
\caption{Performance comparison between CalibraEval and baselines. We report the Fleiss' Kappa (\%) and Intraclass Correlation Coefficient (\%) for each dataset and the averages. The row corresponding to the model name represents the default results without applying any debiasing methods. Best performances are marked bold.}
% \vspace{-4mm}
\begin{tabular}{c|ccc|ccc|ccc|ccc}
% \vspace{-3mm}
\hline
\multirow{2}{*}{Model} & \multicolumn{3}{c|}{RewardBench}                 & \multicolumn{3}{c|}{MTBench}                     & \multicolumn{3}{c|}{PreferenceBench}            & \multicolumn{3}{c}{Average}                      \\
                       & Kappa          & ICC(2,k)       & ICC(3,k)       & Kappa          & ICC(2,k)       & ICC(3,k)       & Kappa          & ICC(2,k)       & ICC(3,k)       & Kappa          & ICC(2,k)       & ICC(3,k)       \\ \hline
Llama-3-8B             & 20.81          & 66.24          & 71.79          & 14.36          & 60.96          & 73.08          & 58.25          & 86.23          & 86.61          & 31.14          & 71.14          & 77.16          \\
DI                     & 19.63          & 64.98          & 70.87          & 15.93          & 59.11          & 65.00          & 39.90          & 76.77          & 80.87          & 25.15          & 66.95          & 72.25          \\
CC                     & 15.49          & 58.77          & 63.70          & 5.60           & 39.48          & 52.45          & 54.84          & 83.60          & 86.26          & 25.31          & 60.62          & 67.47          \\
DC                     & 23.57          & 69.04          & 72.79          & 25.28          & 69.83          & 72.75          & 50.78          & 84.88          & 85.09          & 33.21          & 74.58          & 76.88          \\
Pride                  & 27.65          & 72.72          & 73.77          & 27.38          & 72.27          & 74.50          & 57.01          & 85.49          & 86.33          & 37.35          & 76.83          & 78.20          \\
CalibraEval            & \textbf{30.32} & \textbf{86.51} & \textbf{86.66} & \textbf{28.63} & \textbf{75.45} & \textbf{76.80} & \textbf{58.54} & \textbf{88.17} & \textbf{89.43} & \textbf{39.16} & \textbf{83.38} & \textbf{84.30} \\ \hline
Llama-3.1-8B           & 15.02          & 68.82          & 76.65          & 16.91          & 62.51          & 67.60          & 38.73          & 74.61          & 78.46          & 23.55          & 68.65          & 74.24          \\
DI                     & \textbf{21.59} & 74.04          & 80.00          & 15.12          & 53.24          & 57.08          & 36.59          & 68.47          & 72.14          & 24.43          & 65.25          & 69.74          \\
CC                     & 16.89          & 59.24          & 60.09          & 2.59           & 32.42          & 33.96          & 40.80          & 76.69          & 78.44          & 20.09          & 56.12          & 57.50          \\
DC                     & 23.17          & 72.89          & 76.22          & 16.02          & 65.86          & 68.80          & 41.61          & 77.23          & 78.81          & 26.55          & 71.68          & 74.18          \\
Pride                  & 14.63          & 66.62          & 76.02          & 17.98          & 63.76          & 67.51          & 38.58          & 76.84          & 79.35          & 23.73          & 69.07          & 74.29          \\
CalibraEval            & 20.67          & \textbf{83.23} & \textbf{86.68} & \textbf{19.12} & \textbf{66.00} & \textbf{69.26} & \textbf{43.04} & \textbf{81.56} & \textbf{82.78} & \textbf{27.61} & \textbf{76.93} & \textbf{79.57} \\ \hline
Qwen-14B               & 19.69          & 63.41          & 66.86          & 17.53          & 54.43          & 65.11          & 48.94          & 84.90          & 88.78          & 28.72          & 67.58          & 73.58          \\
DI                     & 11.90          & 50.46          & 53.20          & 4.98           & 36.83          & 46.02          & 30.76          & 74.41          & 82.95          & 15.88          & 53.90          & 60.72          \\
CC                     & -1.62          & -4.71          & -4.93          & -5.48          & 18.08          & 31.20          & 40.35          & 73.79          & 75.56          & 11.08          & 29.05          & 33.94          \\
DC                     & 16.04          & 45.10          & 45.26          & \textbf{25.51} & \textbf{64.37} & \textbf{66.46} & 48.15          & 82.29          & 84.52          & 29.90          & 63.92          & 65.41          \\
Pride                  & 24.73          & 67.46          & 68.30          & 13.00          & 51.73          & 57.39          & 52.79          & 90.86          & 91.20          & 30.17          & 70.02          & 72.30          \\
CalibraEval            & \textbf{26.40} & \textbf{75.75} & \textbf{76.11} & 17.68          & 53.64          & 63.43          & \textbf{62.91} & \textbf{92.56} & \textbf{92.57} & \textbf{35.66} & \textbf{73.98} & \textbf{77.37} \\ \hline
Qwen-72B               & 78.28          & 92.77          & 93.13          & 71.35          & 90.33          & 91.16          & 82.77          & 94.78          & 94.90           & 77.47          & 92.63          & 93.06          \\
DI                     & 77.33          & 92.27          & 92.63          & 68.17          & 89.13          & 89.99          & 83.31          & 95.09          & 95.11          & 76.27          & 92.16          & 92.58          \\
CC                     & 69.23          & 88.71          & 89.92          & 70.64          & 90.29          & 90.72          & 78.08          & 92.63          & 93.24          & 72.65          & 90.54          & 91.29          \\
DC                     & 66.61          & 87.03          & 87.90          & 64.59          & 87.46          & 88.45          & 74.03          & 91.01          & 92.01          & 68.41          & 88.50          & 89.45          \\
Pride                  & 78.64          & 92.99          & 93.30          & 71.50          & 90.54          & 91.27          & 83.44          & 94.89          & 95.00          & 77.86          & 92.81          & 93.19          \\
CalibraEval            & \textbf{82.80} & \textbf{95.47} & \textbf{95.75} & \textbf{71.88} & \textbf{95.70} & \textbf{96.71} & \textbf{85.25} & \textbf{97.56} & \textbf{97.57} & \textbf{79.98} & \textbf{96.24} & \textbf{96.68} \\ \hline
ChatGPT                & 20.08          & 62.67          & 70.75          & 37.25          & 73.90          & 76.92          & 64.62          & 87.63          & 87.81          & 40.65          & 74.73          & 78.49          \\
DI                     & 24.52          & 70.23          & 71.58          & 24.33          & 66.80           & 67.69          & 56.00          & 82.90          & 84.89          & 34.95          & 73.31          & 74.72          \\
CC                     & 24.28          & 57.25          & 58.23          & 23.71          & 64.06          & 71.91          & 61.63          & 86.18          & 86.38          & 36.54          & 69.16          & 72.17          \\
DC                     & 27.94          & 66.09          & 70.54          & 16.68          & 58.37          & 70.26          & 55.33          & 81.92          & 82.58          & 33.32          & 68.79          & 74.46          \\
Pride                  & 28.25          & 70.38          & 73.16          & 39.02          & 76.61          & 77.61          & 64.64          & 87.56          & 87.82          & 43.97          & 78.18          & 79.53          \\
CalibraEval            & \textbf{32.02} & \textbf{77.25} & \textbf{77.60} & \textbf{39.71} & \textbf{79.07} & \textbf{79.85} & \textbf{65.52} & \textbf{87.77} & \textbf{87.92} & \textbf{45.75} & \textbf{81.36} & \textbf{81.79} \\ \hline
GPT4o                  & 82.57          & 94.83          & 94.89          & 72.42          & 92.99          & 93.20          & 79.42          & 93.50          & 94.11          & 78.14          & 93.77          & 94.07          \\
DI                     & 78.57          & 93.72          & 94.01          & \textbf{74.21}          & 93.35          & 93.50          & \textbf{79.97}          & 94.48          & 94.93          & 77.58          & 93.85          & 94.15          \\
CC                     & 81.38          & 94.47          & 94.48          & 67.10          & 90.30          & 90.78          & 77.56          & 92.76          & 93.43          & 75.35          & 92.51          & 92.90          \\
DC                     & 76.89          & 92.28          & 92.35          & 68.94          & 90.94          & 91.64          & 70.48          & 89.30          & 90.02          & 72.10          & 90.84          & 91.34          \\
Pride                  & 82.53          & 94.94          & 94.98          & 70.34          & 92.35          & 92.74          & 79.74          & 93.62          & 94.20          & 77.54          & 93.64          & 93.97          \\
CalibraEval            & \textbf{83.25} & \textbf{96.25} & \textbf{96.27} & 72.60 & \textbf{95.04} & \textbf{95.20} & 79.73 & \textbf{97.29} & \textbf{97.60} & \textbf{78.53} & \textbf{96.19} & \textbf{96.36} \\ \hline
\end{tabular}

\label{main}
% \vspace{-5mm}
\end{table*}

\subsection{Baselines}
We employ the following methods as our baselines. Since CalibraEval is a label-free method, we do not compare it with supervised methods.
\begin{itemize}[leftmargin=*]
    \item \textbf{Debiasing Instruct (DI)} is implemented by including the instruction: ``Avoid any position bias and ensure that the order in which the responses were presented does not influence your decision. Do not allow the length of the responses to influence your evaluation. Do not favor certain tokens of the option. Be as objective as possible''.
    \item \textbf{Contextual Calibration (CC)}~\cite{zhao2021calibrate} involves applying an affine transformation to model outputs in order to calibrate LLM predictions. It estimates the bias for each option tokens by requesting its prediction with a prompt alongside a content-free input, such as ``N/A''.
    \item \textbf{Domain-context Calibration (DC)}~\cite{fei2023mitigating} is designed to minimize label bias in in-context learning. It estimates a contextual prior by using a random in-domain sequence, achieving state-of-the-art results.
    \item \textbf{Pride}~\cite{zheng2023large} estimates the model's prior bias toward option ID token by reorganizing the test samples and then removes this bias using a division operation.
\end{itemize}

% 如表3所示，我们进一步分析了 ClaibraEval 的性能，使用Label相关的指标。受限于空间，我们只汇报了Qwen-72B，ChatGPT、GPT4o上的实验结果，完整的结果见附录A。在表3中，我们汇报了3种组合下的Rstd和Acc的平均。
% 我们发现，CalibraEval有着更小的Rstd表现，这表明它能够更有效的降低选择误差。尽管我们的方法设计的目的并不是为了增加准确率，但我们发现，往往更小的Rstd往往意味着更高的准确率，且去偏效果越明显时，性能增加越多。例如ChatGPT在RewardBench上Rstd从16.79降至5.51，准确率从65.27提升到67.13。这可能说明选择误差一定程度影响到模型的判断，更好的去偏能导致更有效的评测结果。 

\subsection{Implementation Details}
We evaluate six models from three LLM-families including: Llama-3-8B~\cite{touvron2023llama}, Llama-3.1-8B~\cite{touvron2023llama}, Qwen-14B~\cite{bai2023qwen}, Qwen-72B~\cite{bai2023qwen}, ChatGPT~\cite{openai2023gpt4}, and GPT-4o~\cite{openai2023gpt4}. The version of ChatGPT used is gpt-3.5-turbo-1106.
The estimation set used to derive the calibration function can be constructed either by sampling from the test data or by using the entire test set without the gold labels. For a fair comparison, we opted for the latter approach. In the main experiment, all baselines used the full test data as the prior estimation set.
For CC, we use the predefined token ``N/A'' to replace the option contents, generating content-free input. For DC, we randomly extract words from the task corpus to construct the content-free input.
Moreover, we set $\lambda=0.5$ and $\gamma=10$. We employ the batch gradient descent method with a batch size of 32. The optimization process stops when the parameters change range is less than the threshold $\epsilon$ i.e., 
$\sum_{i=1}^{N}\bigtriangleup d_i < \epsilon$. The $\epsilon$ is set to 0.001. All experiments presented in this paper are conducted on 8 NVIDIA Tesla A100 GPUs. All the prompts used in this paper can be found in Appendix \ref{prompt}.

\section{Experiment Result}
\subsection{Main Results}
To validate the effectiveness of CalibraEval in mitigating selection bias, we test the consistency of evaluation results among different models on benchmarks. The performance comparison of CalibraEval with baselines is presented in Table \ref{main}. Based on the experimental results, we can draw the following conclusions.

\begin{itemize}[leftmargin=*]
    \item Debiasing Instruct does not consistently lead to improved or more robust performance, as its effectiveness is limited by the instruction-following capabilities of LLMs and the nature of tasks. In some cases, adding debiasing instructions may even result in consistency degradation. Consequently, relying solely on instructions is not a reliable approach for effective debiasing.
    \item CC and DC are originally designed to mitigate label bias in in-context learning. Therefore, their estimated priors may not accurately reflect the inherent selection bias in LLMs-as-Judges, leading to suboptimal debiasing performance and difficulties in interpretation.
    \item When applied to lower-capability LLMs, such as Llama-3-8B and Qwen-14B, Pride effectively estimates bias and improves consistency. However, its effectiveness diminishes with more advanced models (e.g., GPT4o). This limitation may arise from the simplified probabilistic relationships employed in Pride.        
    \item CalibraEval consistently improves performance across various LLMs and tasks. On average, CalibraEval shows enhancements over all the baselines. 
    Overall, CalibraEval is a versatile technique applicable to multiple evaluation tasks, delivering stable performance improvement.
    This also indicates that CalibraEval can effectively reduce selection bias in LLMs-as-judges, leading to more consistent and fair evaluation results.    
\end{itemize}

Table \ref{acc} presents the performance of the reference-based metrics.
Due to space constraints, we only report the experimental results for Llama-3-8B, Qwen-14B, and ChatGPT, while the complete results are available in Appendix \ref{more_results}. 
For a fair comparison, we report the average values of Rstd and Accuracy under the conditions of swapping option positions and option IDs.
Across the average performance of the three datasets, CalibraEval consistently achieves lower Rstd and higher accuracy, outperforming other baselines.
Surprisingly, although this is not the original intent, CalibraEval frequently improves accuracy.
We believe this may indicate that selection bias influences the model's judgments, leading to reduced accuracy. Therefore, effective bias mitigation methods can enhance the model's performance in its evaluative role.
Additionally, we found that lower Rstd is often associated with higher accuracy. The more pronounced the debiasing effect, the more significant the performance improvement. For example, on RewardBench, ChatGPT's Rstd decreased from 16.79 to 5.51, while its accuracy increased from 65.27 to 67.13. Overall, CalibraEval not only enhances the reliability of model evaluations but also unlocks the potential for these LLMs to perform optimally in various tasks.

\begin{table}[]
\caption{
Results of reference-based metrics. We report the Standard Deviation of Recalls (RStd) and Accuracy (Acc.), with the best results highlighted in bold. $\downarrow$ indicates that lower values correspond to better performance.}
\begin{tabular}{c|cc|cc|cc}
\hline
\multirow{2}{*}{Model} & \multicolumn{2}{c|}{RewardBench} & \multicolumn{2}{c|}{MTbench}   & \multicolumn{2}{c}{Preference Bench} \\
                       & Rstd           & Acc.(\%)        & Rstd          & Acc.(\%)       & Rstd             & Acc.(\%)          \\ \hline
Llama-3-8B             & 15.01          & 65.79           & 16.42         & 67.08          & 3.36             & 83.43             \\
Pride                  & 7.51           & 66.54           & 11.64         & 70.63          & 4.35             & 83.24             \\
CalibraEval            & \textbf{6.48}  & \textbf{68.12}  & \textbf{5.22} & \textbf{70.63} & \textbf{3.42}    & \textbf{83.98}    \\ \hline
Qwen-14B               & 11.63          & 63.14           & 17.24         & 65.61          & 11.99            & 80.68             \\
Pride                  & 4.18           & 64.09           & 16.31         & 65.29          & 7.36             & 83.55             \\
CalibraEval            & \textbf{2.72}  & \textbf{64.25}  & \textbf{6.26} & \textbf{68.64} & \textbf{5.12}    & \textbf{83.88}    \\ \hline
ChatGPT                & 16.79          & 65.27           & 7.66          & 72.67          & 3.04             & 85.61             \\
Pride                  & 8.54           & 66.36           & 6.01          & 72.86          & 3.51             & 85.68             \\
CalibraEval            & \textbf{5.51}  & \textbf{67.13}  & \textbf{5.20} & \textbf{72.98} & \textbf{2.82}    & \textbf{85.98}    \\ \hline
\end{tabular}
\label{acc}
\end{table}

\begin{figure}[t]
    \centering
    \begin{subfigure}[ChatGPT-ICC]{
        \centering
        \includegraphics[width=0.45\linewidth]{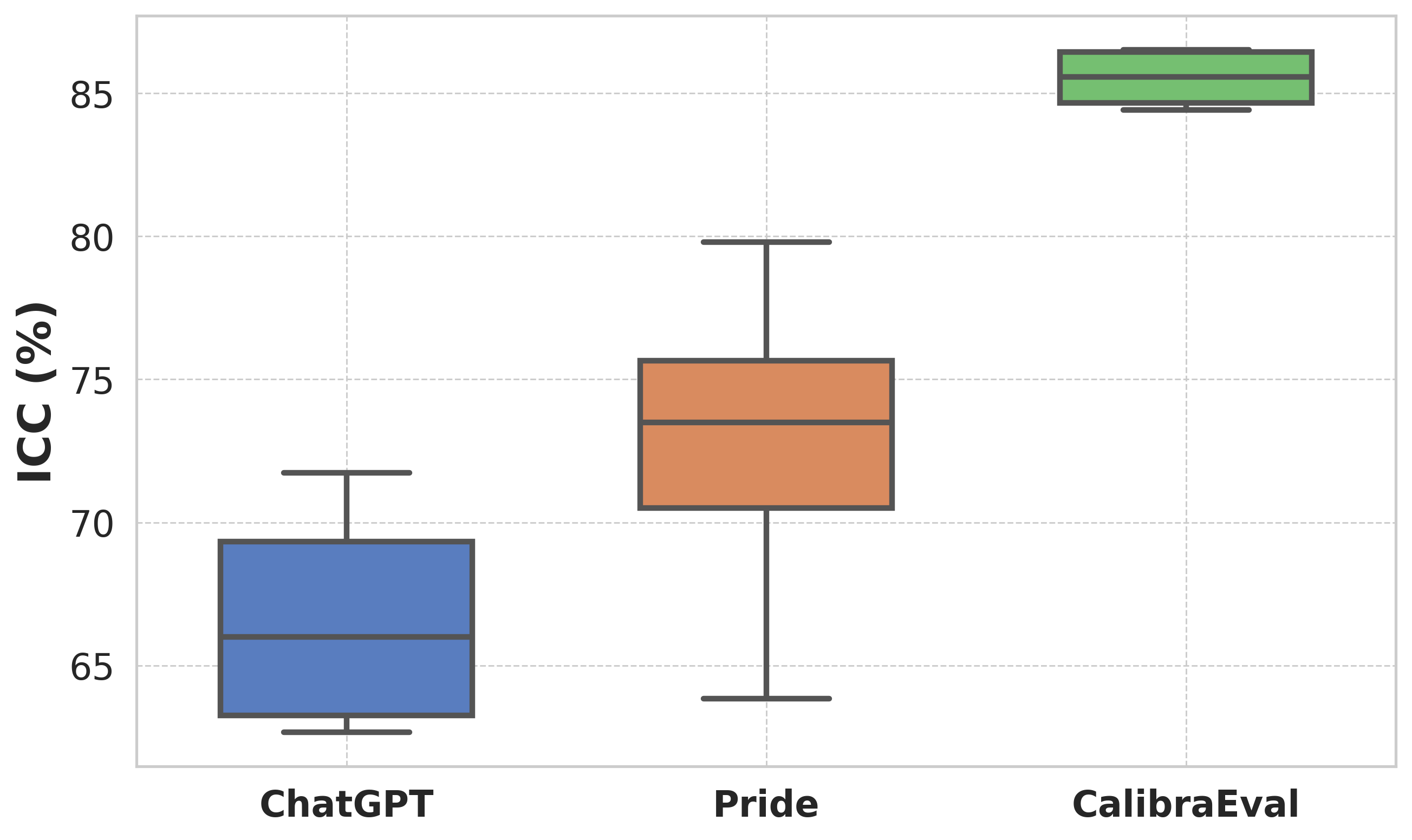}
        }
    \end{subfigure}
    \begin{subfigure}[ChatGPT-Kappa]{
        \centering
        \includegraphics[width=0.45\linewidth]{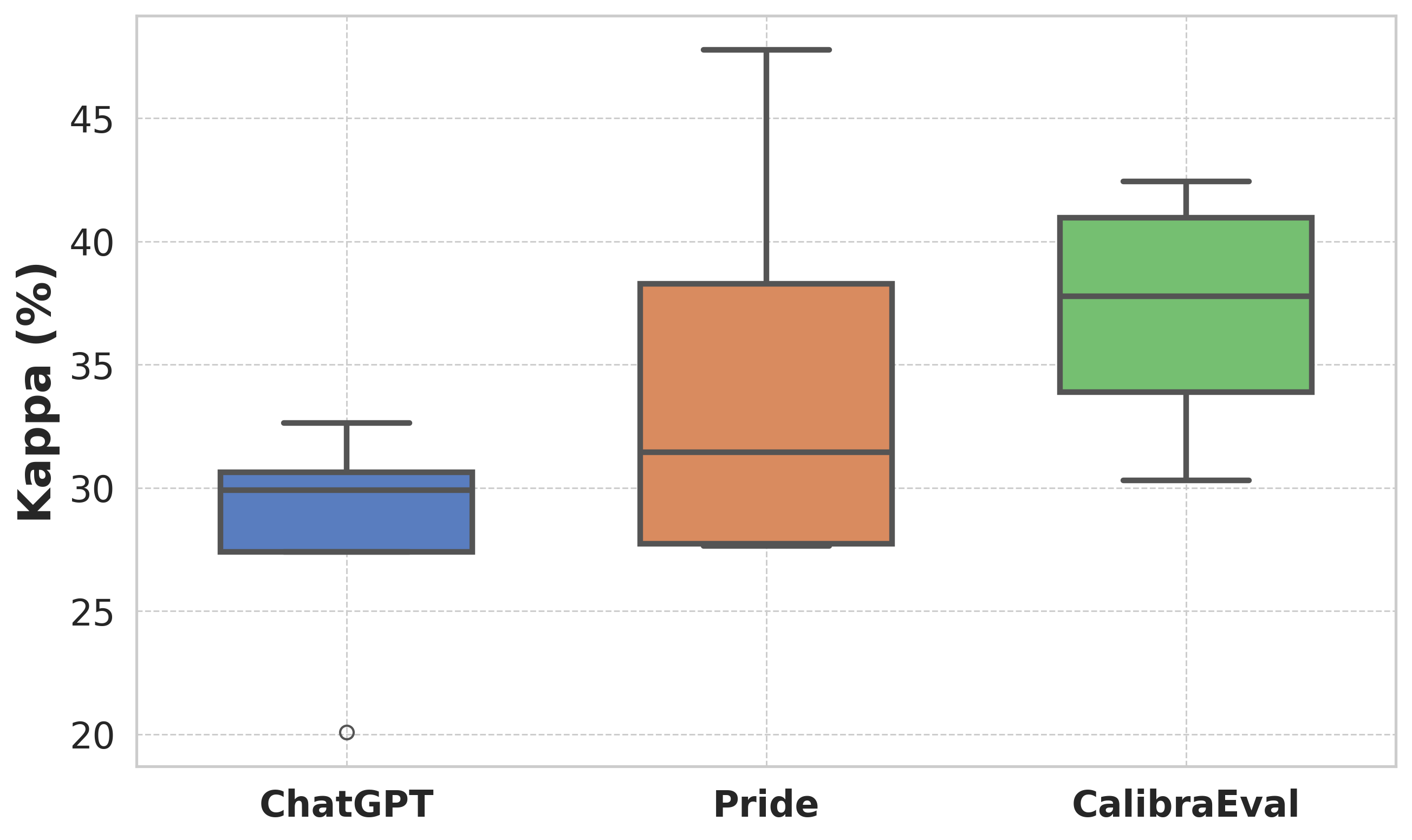}
        }
       
    \end{subfigure}

    \begin{subfigure}[Qwen72B-ICC]{
        \centering
        \includegraphics[width=0.45\linewidth]{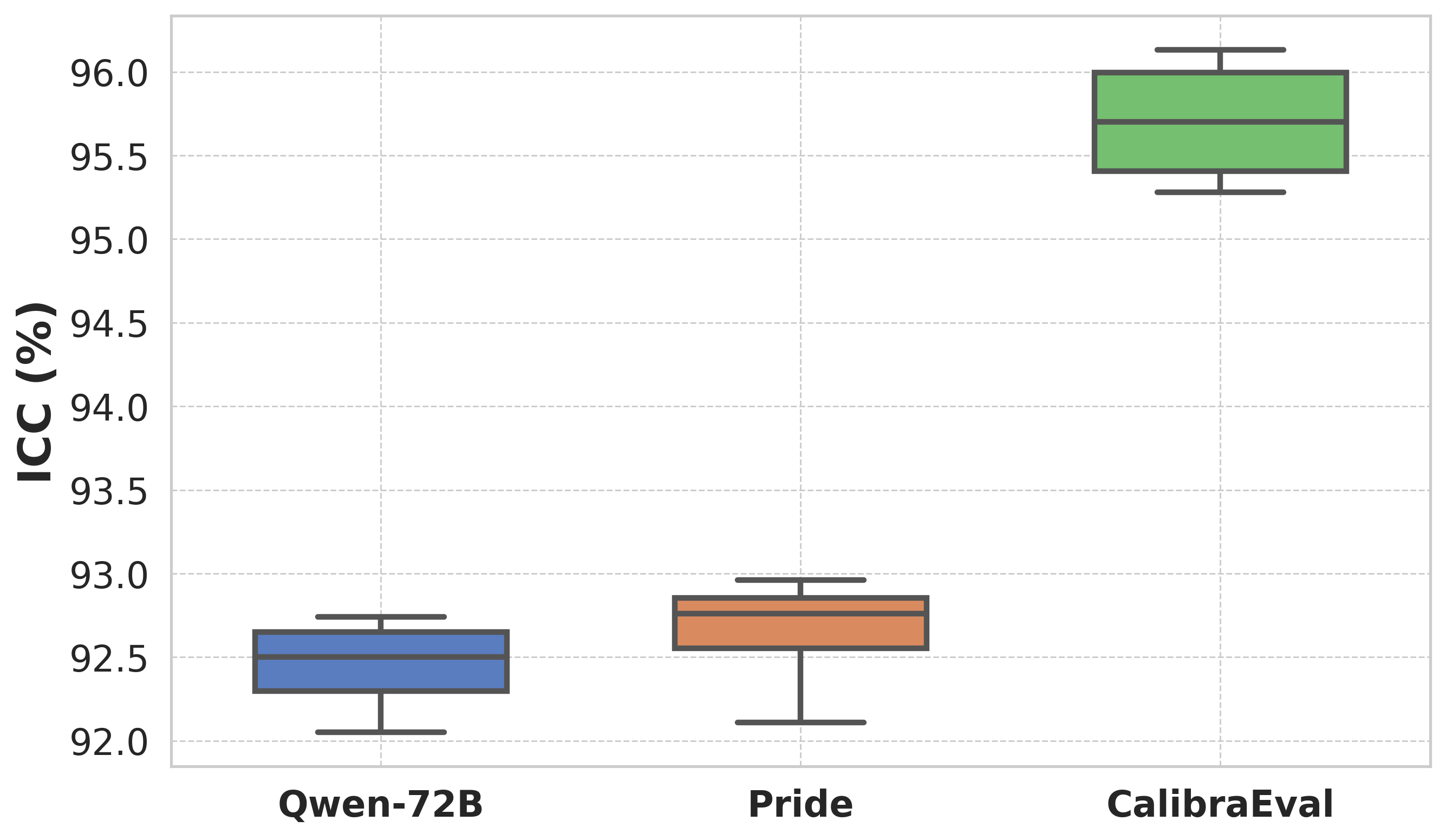}
        }
    \end{subfigure}
    \begin{subfigure}[Qwen72B-Kappa]{
        \centering
        \includegraphics[width=0.45\linewidth]{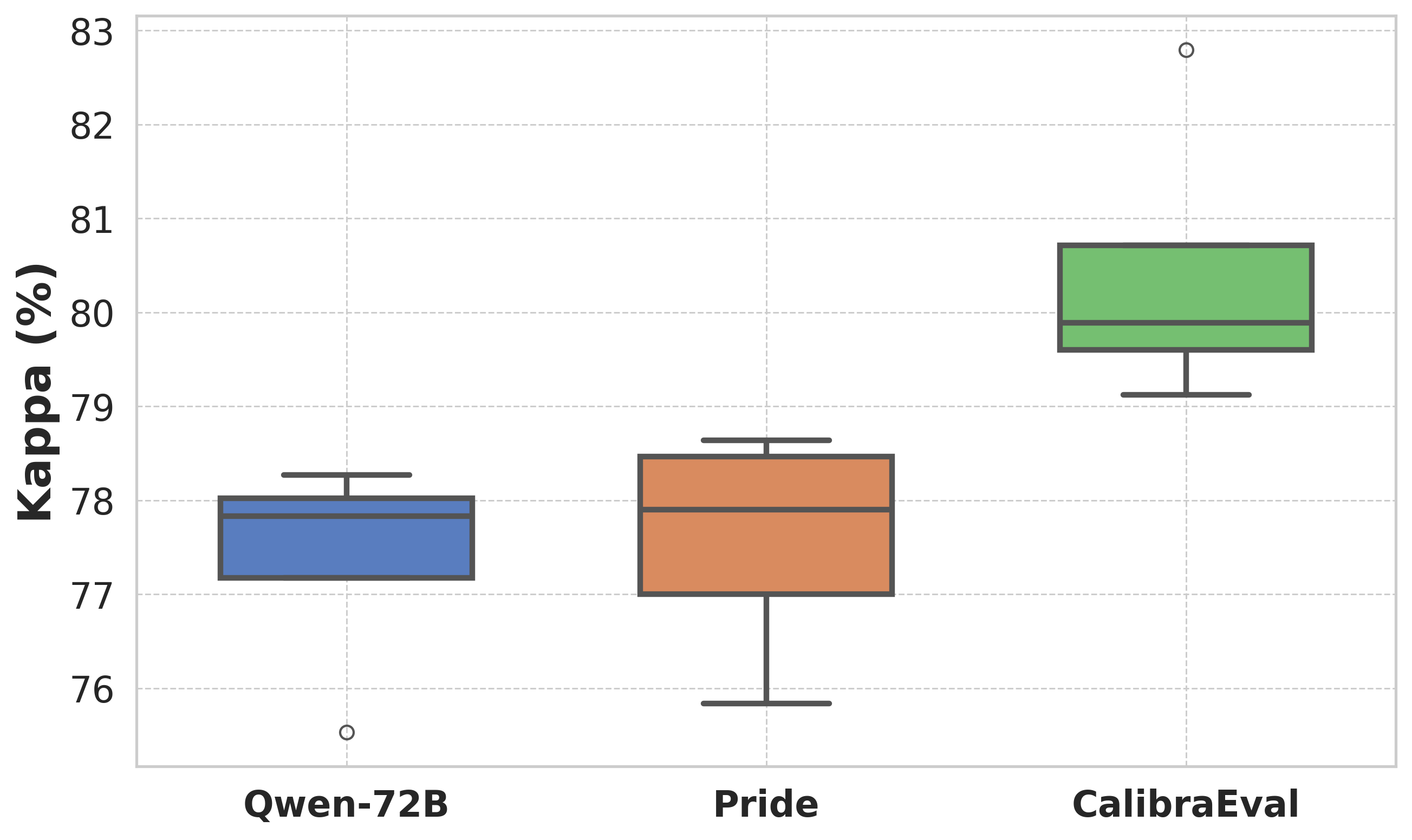}
        }
    \end{subfigure}
    
    \caption{Performance comparison across different prompt templates.}
    \label{prompt}
    % \vspace{-5mm}
\end{figure}

\begin{figure}[t]
    \centering
    \label{tokens}
    \begin{subfigure}[ChatGPT-ICC]{
        \centering
        \includegraphics[width=0.45\linewidth]{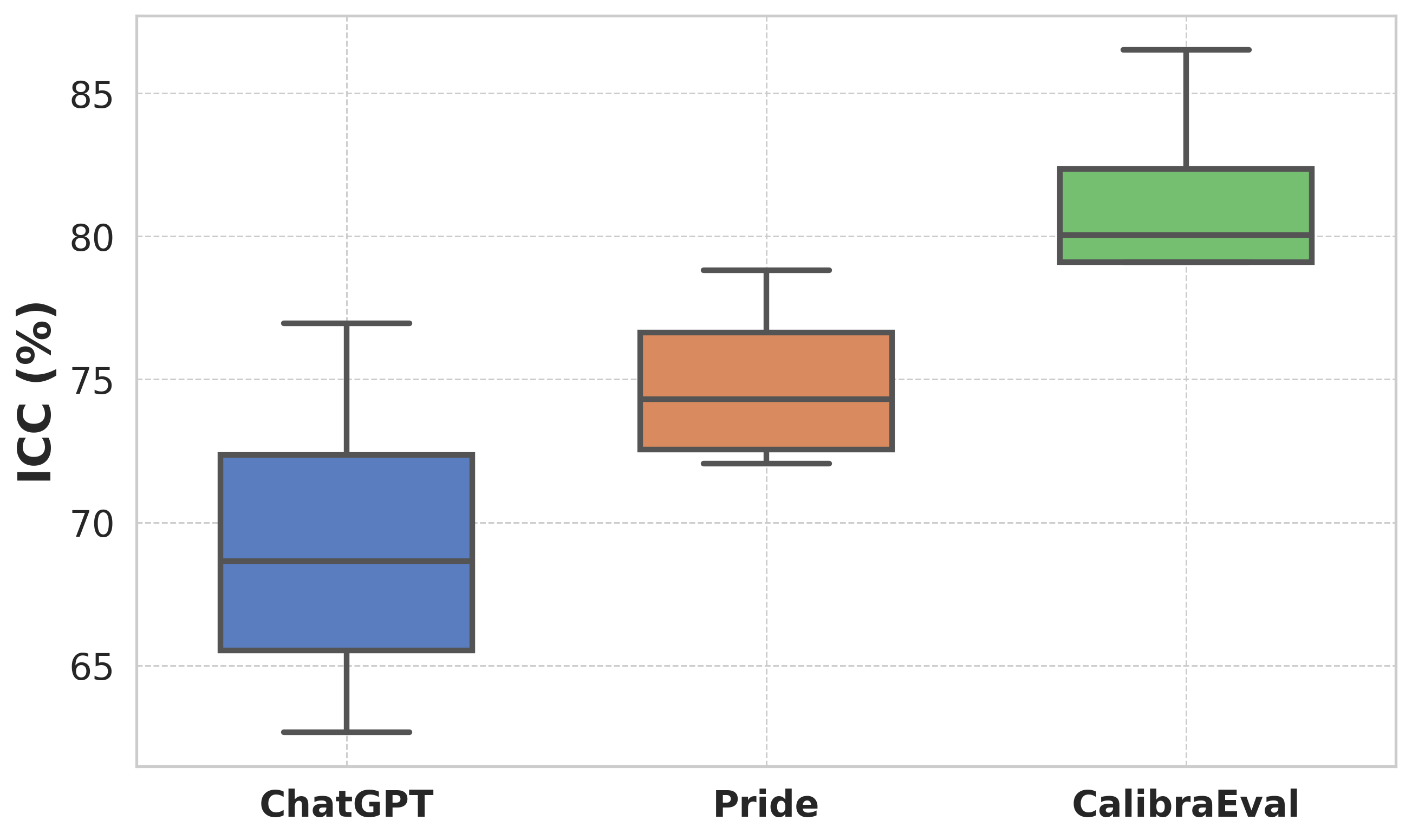}
        }
    \end{subfigure}
    \begin{subfigure}[ChatGPT-Kappa]{
        \centering
        \includegraphics[width=0.45\linewidth]{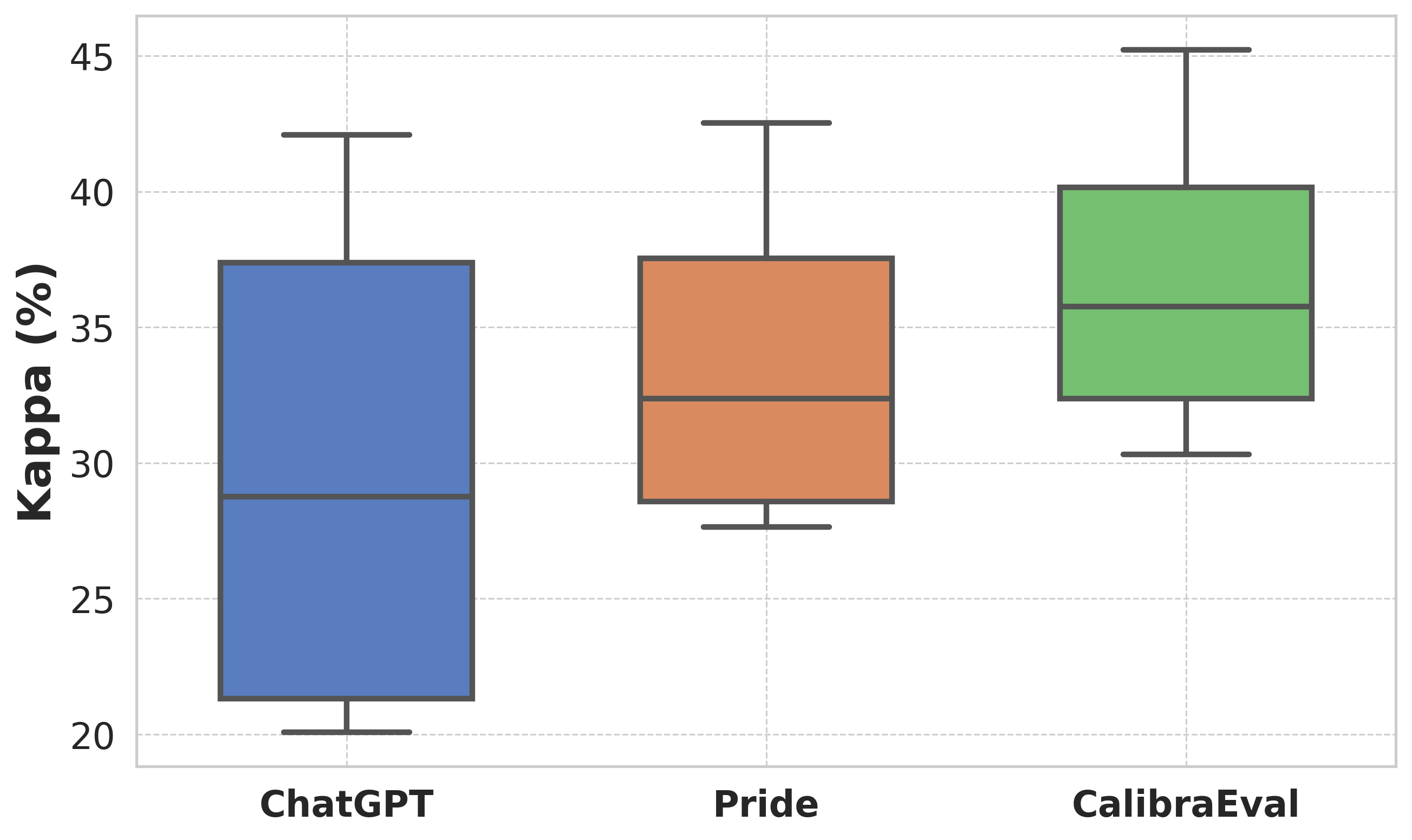}
        }
       
    \end{subfigure}

    \begin{subfigure}[Qwen72B-ICC]{
        \centering
        \includegraphics[width=0.45\linewidth]{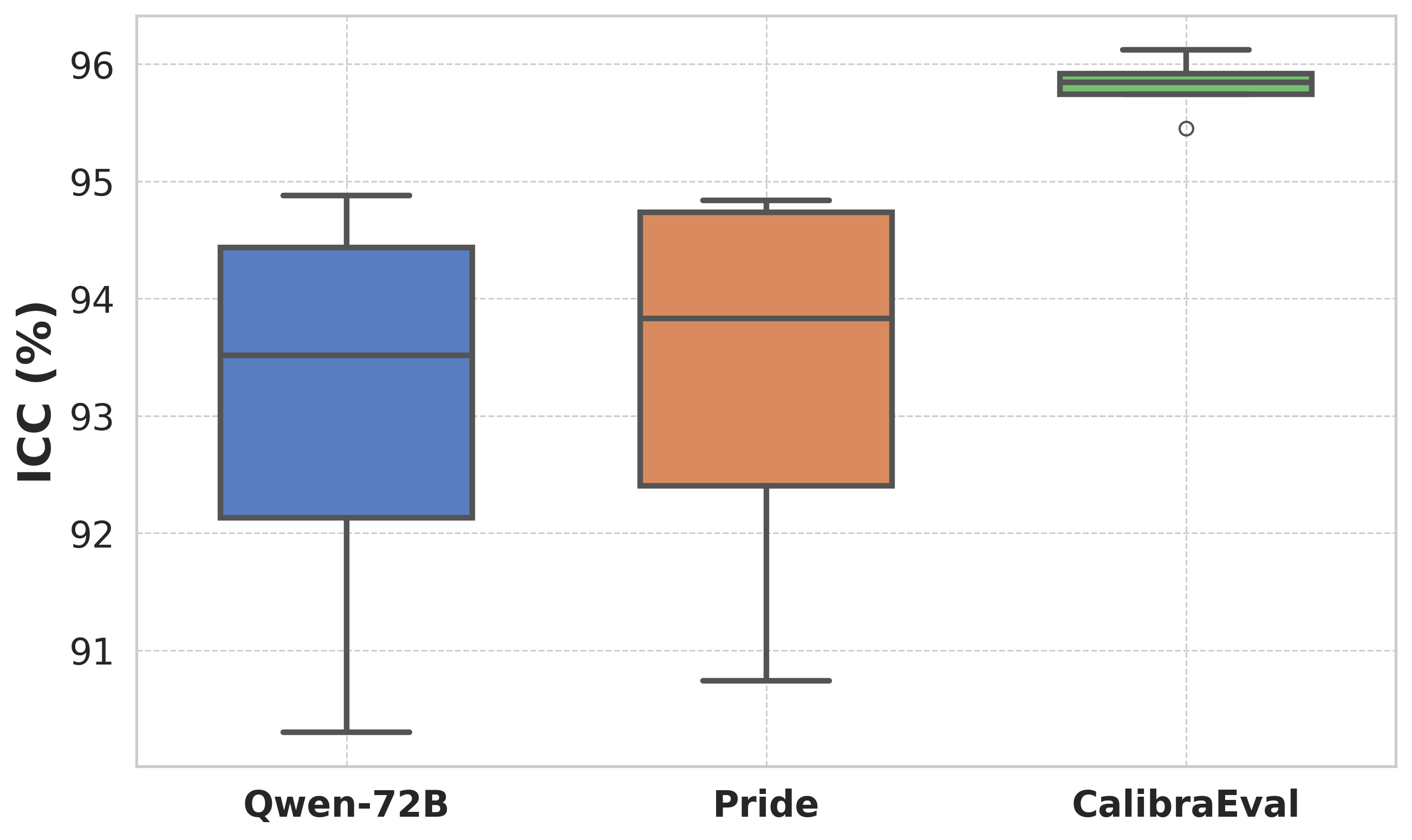}
        }
    \end{subfigure}
    \begin{subfigure}[Qwen72B-Kappa]{
        \centering
        \includegraphics[width=0.45\linewidth]{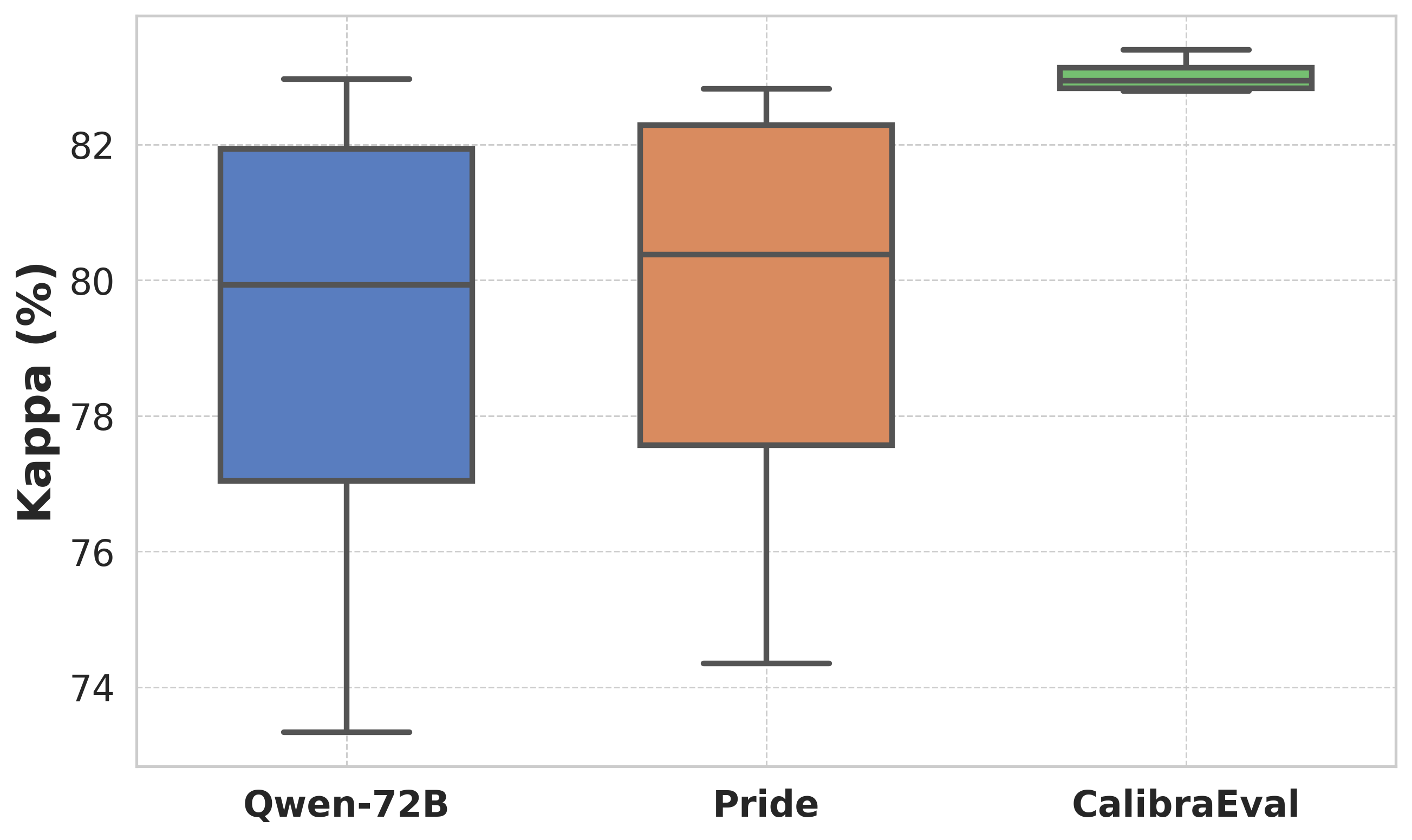}
        }
    \end{subfigure}

    \caption{Performance comparison across different ID tokens.}
    % \vspace{-5mm}
    \label{tokens}

\end{figure}

\subsection{Robustness Analysis}
\label{Robustness}
In this section, we conduct additional experiments to further validate the effectiveness of CalibraEval across diverse scenarios. Due to the high cost of GPT-4o, we opt for Qwen-72B and ChatGPT on the RewardBench for the following experiments. Unless otherwise stated, the ICC for subsequent experiments is ICC(2,k).

\subsubsection{Different prompt
templates}

We conduct experiments on four distinct prompt templates (see Appendix \ref{prompt} for details). Figure \ref{prompt} shows performance comparisons on RewardBench. We observed that model outputs without bias correction exhibit low consistency and high variance. While Pride improves consistency, it still exhibited considerable variance. In contrast, CalibreEval demonstrates substantial performance enhancement while maintaining low variance, indicating its consistent effectiveness across different prompt templates.

\subsubsection{Different ID
tokens}

We also conduct experiments using four distinct sets of ID tokens: A/B, a/b, Alice/Bob, and X/Y. Figure~\ref{tokens} illustrates the performance comparison. 
CalibraEval consistently achieves significant performance improvements with low variance across all tested ID tokens. This highlights its robustness and effectiveness regardless of the specific tokens used. Furthermore, when applied to the highly consistent model Qwen-72B, the improvement of Pride is negligible, while CalibreEval continued to enhance consistency even further.

\subsubsection{Different number of in-context learning examples}

\begin{figure}[t]
    \centering
    \begin{subfigure}[ChatGPT]{
        \centering
        \includegraphics[width=0.8\linewidth]{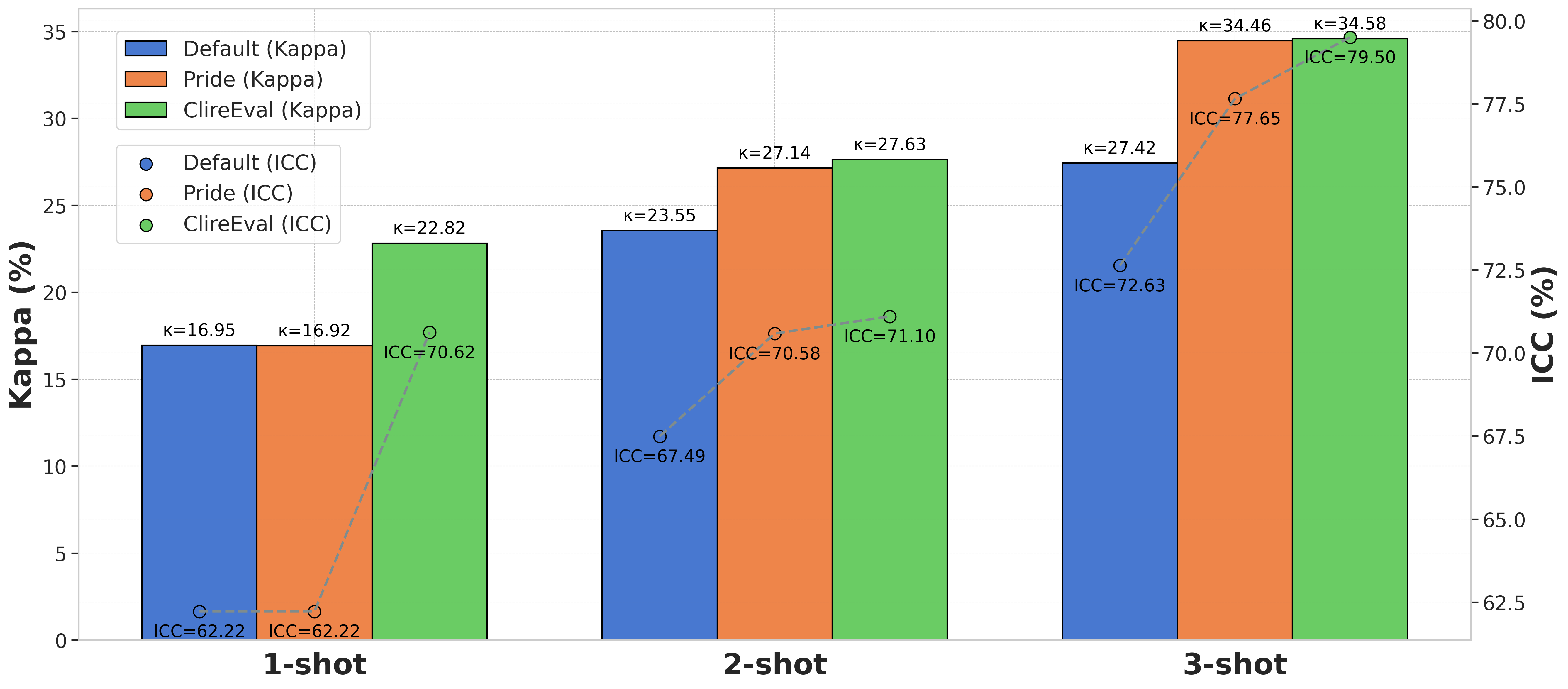}
        }
    \end{subfigure}
    \begin{subfigure}[Qwen-72B]{
        \centering
        \includegraphics[width=0.8\linewidth]{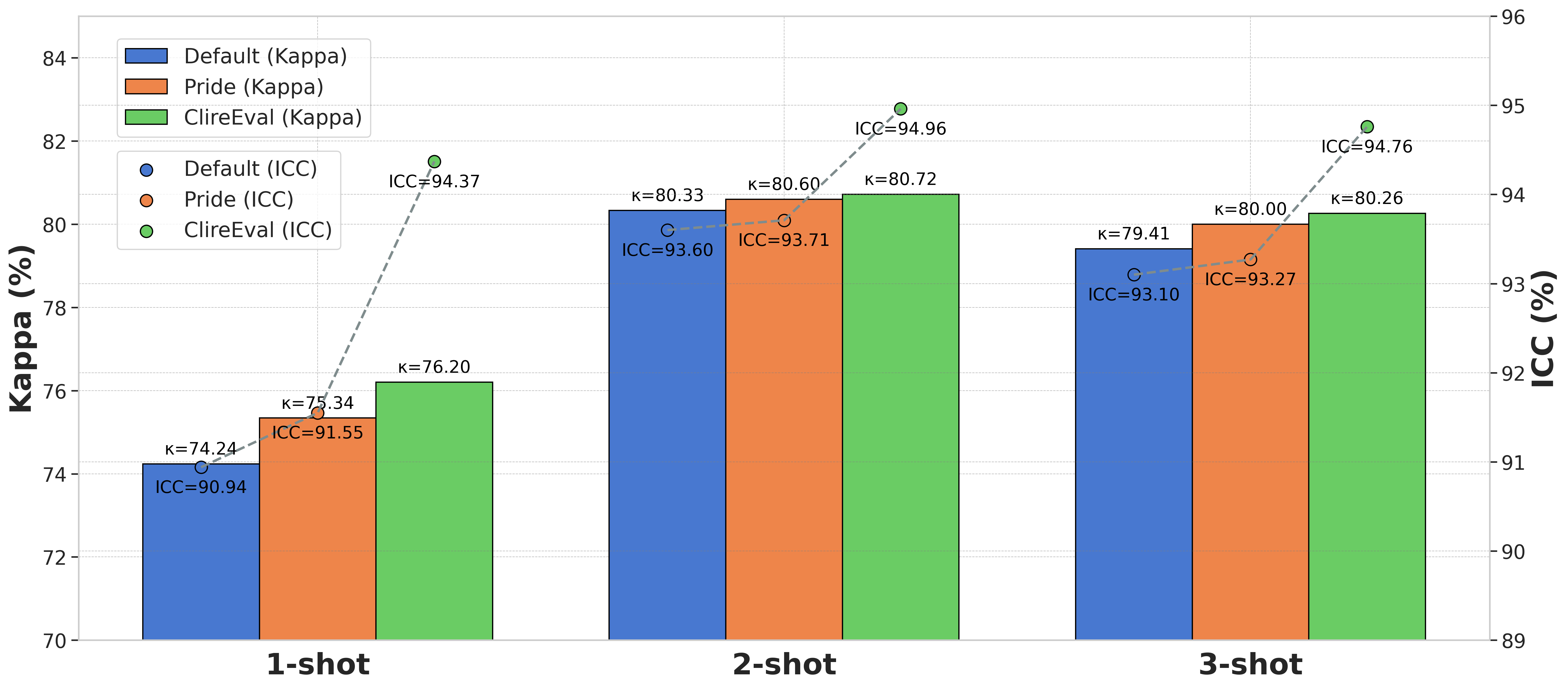}
        }
       
    \end{subfigure}
% \vspace{-5mm}

\caption{Performance comparison under in-context learning.}
% \vspace{-5mm}

\label{few}
\end{figure}

We further investigate the effectiveness of CalibraEval when conducting in-context learning. Specifically, we provided the LLMs with 1-shot, 2-shot, and 3-shot examples, respectively. 
As shown in Figure \ref{few}, CalibraEval remains effective even when examples are provided for in-context learning. 
We find that as the number of examples increases, the consistency of the model's judgments also improves. This may be because examples help the model better understand the task, leading to more confident and consistent evaluations. 
Additionally, we also observed that the effectiveness of calibration methods like Pride and CalibraEval decreases as the number of examples increases.
This may be due to these examples introducing new biases, which affect the effectiveness of the calibration.
Therefore, we believe that calibration methods have greater potential for application in zero-shot scenarios.

% The above experiments demonstrate the effectiveness of CalibraEval in various scenarios, showcasing the broad applicability of our approach.

% .然而，PROCA明显优于其他排列，并且在不同的排列中保持着极低的方差，表明其对类比例和排列的不敏感性。
% 我们进行了更多实验来验证 ClibraEval的有效性。
% 由于GPT4o成本过高，我们选择Qwen72B和ChatGPT两个模型在RewardBench上开展下述实验。

% \textbf{PROCA 在不同的模板中都能持续有效。我们在九种不同的提示模板和标签空间中进行了实验（模板详情见附录表 8）。三种方法在 SST-2 上的性能比较如图 4 所示。我们发现，上下文校准虽然提高了平均准确率，但仍存在较大的方差。然而，我们提出的原型校准法却能以较低的方差带来较大的改进，这表明 PROCA 对各种提示模板都很有效。}

%  ClibraEval在不同的option ID tokens持续有效

%  CLibraEval 在Few-shot上持续有效。

\begin{figure}[t]
    \centering
    \begin{subfigure}[ChatGPT]{
        \centering
        \includegraphics[width=0.7\linewidth]{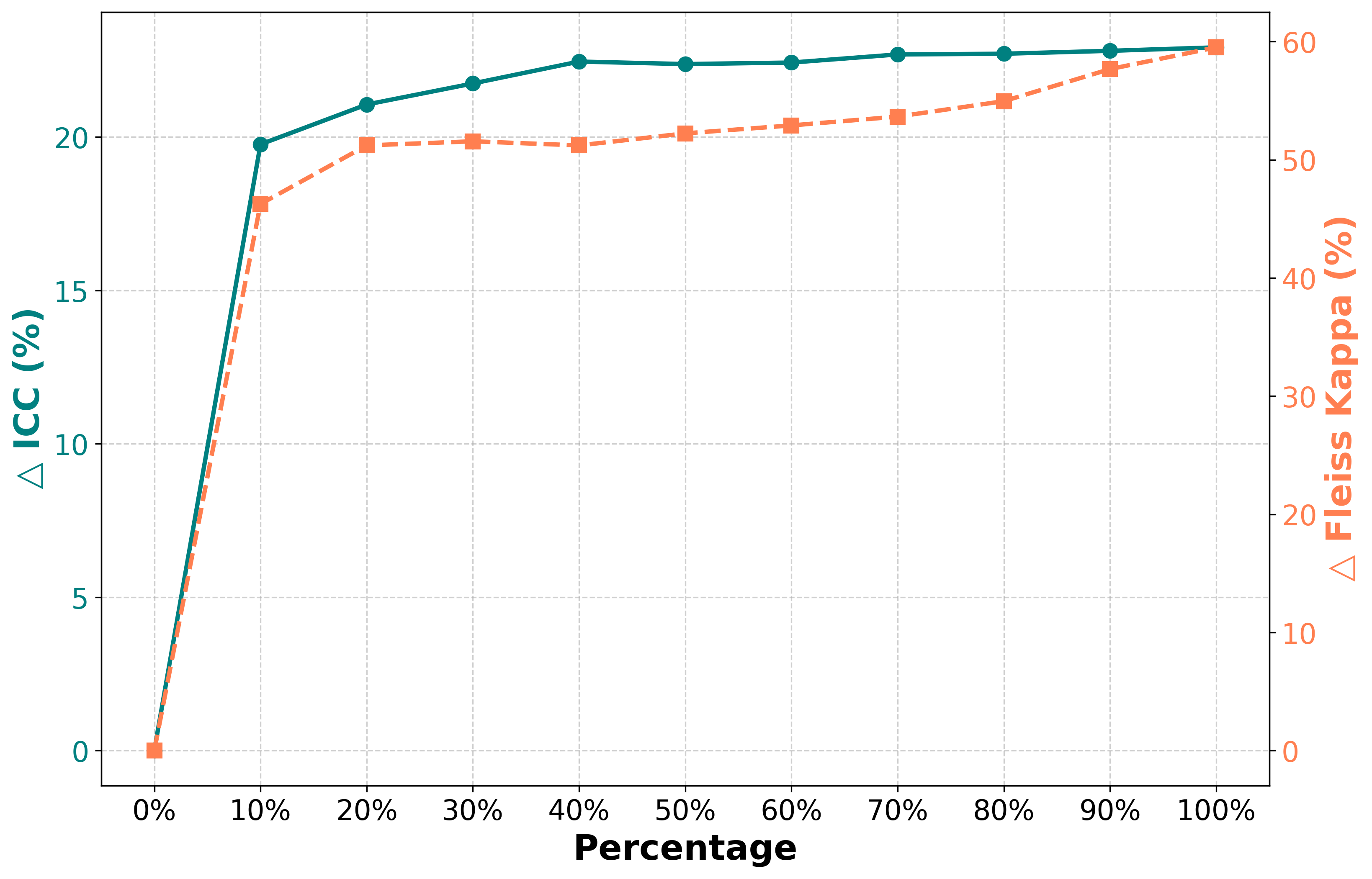}
        }
    \end{subfigure}
    \begin{subfigure}[Qwen-72B]{
        \centering
        \includegraphics[width=0.7\linewidth]{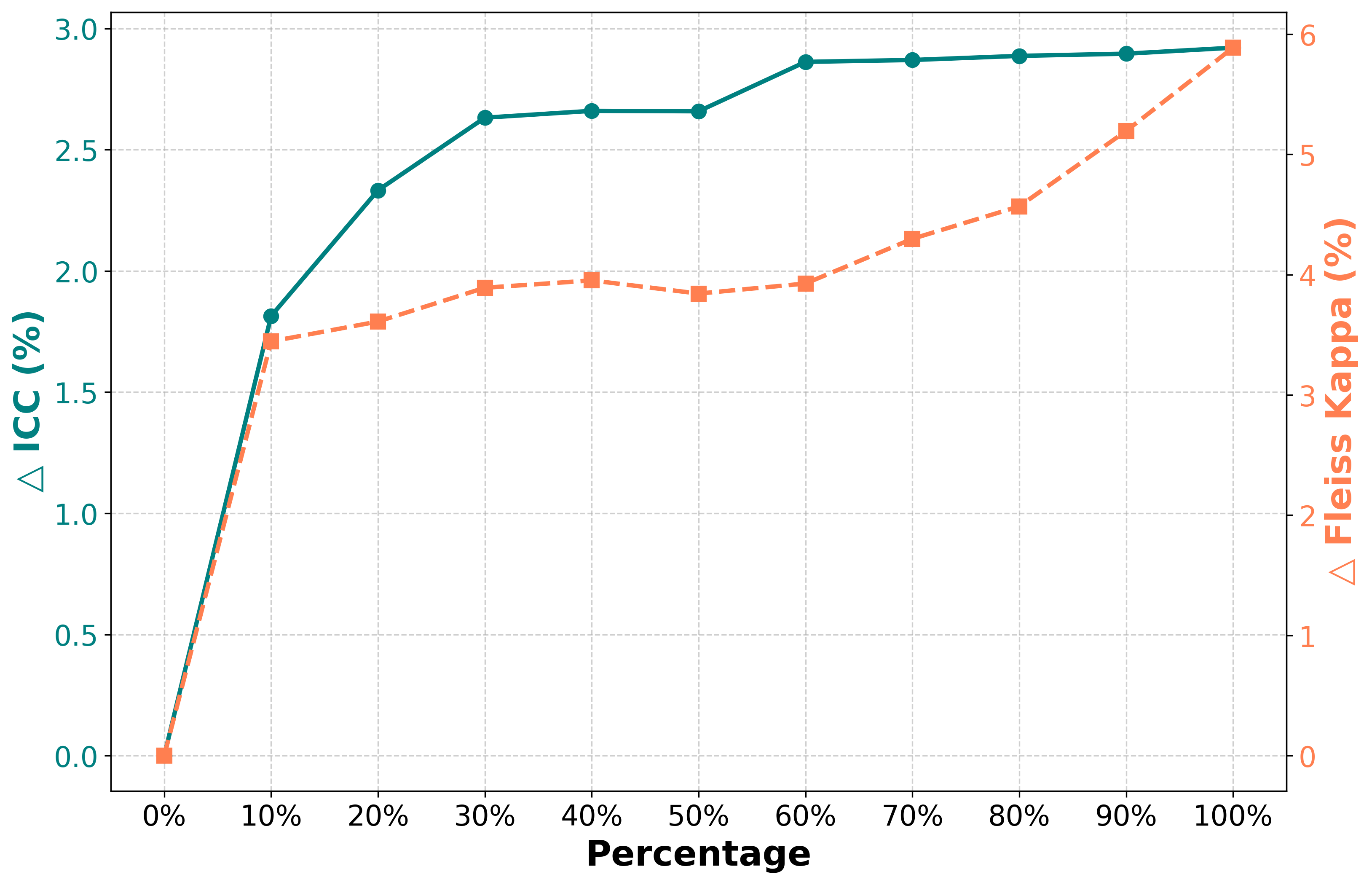}
        }
       
    \end{subfigure}
    % \vspace{-5mm}

\caption{ Performance of CalibraEval across different estimate set sizes. ``Percentage'' refers to the proportion of the test set selected for use as the estimation set.}
    % \vspace{-5mm}

\label{setsize}
\end{figure}

\subsection{Ablation Studies}
To better illustrate the rationality and effectiveness of model design, we conduct two ablation experiments.
We first analyze the effectiveness of well-defined optimization objectives.
Specifically, we consider two variants.  The first variant focuses solely on ensuring that the model maintains consistent judgments after swap ID tokens, i.e., 
\begin{equation} \label{eqn-218}
L_1 =  arg\min_{g\in G}  \sum_{i=1}^{K}[ g(s_{0}^{i}) + g(s_2^i)-1]^2 -\lambda[ g(s_{0}^{i}) - g(s_2^i)]^2
\end{equation}

The other variant focuses on ensuring that the model maintains consistent judgments after position exchanges, represented as:  
\begin{equation} \label{eqn-219}
L_2 =  arg\min_{g\in G}  \sum_{i=1}^{K}[ g(s_{0}^{i}) - g(s_1^i)]^2 -\lambda[ g(s_{0}^{i}) - 0.5]^2
\end{equation}
Since this variant does not involve $s_2^i$, the regularization term is modified to $[g(s_{0}^{i}) - 0.5]^2$ to prevent the model from converging to a trivial solution.

Table \ref{abla} illustrates the impact of different optimization objectives. 
Both objectives contribute to the calibration benefits observed.
When the model is significantly influenced by position bias, the improvements from $L_2$ are more substantial. Conversely, when token bias is more prevalent, $L_1$ leads to better improvements. The combination of both objectives, which defines our CalibraEval optimization goal, achieves optimal performance. These experiments validate the effectiveness of our chosen optimization settings.

In Figure \ref{setsize}, we further test the impact of the estimation set size on the performance.
We randomly sampled a certain proportion of test data to estimate the calibration function, which is then applied to debias the entire test set. 
We found that increasing the size of the estimation set can better enhance consistency. Additionally, a smaller estimation set can also effectively support CalibraEval in reducing bias. For ChatGPT, using only 10\% of the data resulted in improvements of over 85\% compared to the full dataset.  
Overall, even with limited data, CalibraEval can still produce reliable calibration functions.

\begin{table}[t]
\caption{Ablation study on RewardBench. Best results are marked bold.}
% \vspace{-4mm}
\begin{tabular}{c|ccccc}
\hline
Model       & Kappa          & ICC(2,k)       & ICC(3,k)       & Rstd$\downarrow$          & Acc.           \\ \hline
ChatGPT     & 20.08          & 62.67          & 70.75          & 16.79         & 65.27          \\
w. $L_1$& 26.78          & 72.04          & 73.66          & 10.32         & 65.33          \\
w. $L_2$ & 27.38          & 72.73          & 75.99          & 8.64          & 66.04          \\
w. both      & \textbf{32.02} & \textbf{77.25} & \textbf{77.60} & \textbf{5.50} & \textbf{67.13} \\ \hline
Qwen-72B    & 78.28          & 92.77          & 93.13          & 4.01          & 87.20          \\
w. $L_1$ & 81.95          & 94.94          & 95.04          & 2.42          & 87.78          \\
w. $L_2$ & 81.32          & 93.77          & 95.52          & 2.78          & 87.74          \\
w. both      & \textbf{82.80}  & \textbf{95.47} & \textbf{95.75} & \textbf{0.94} & \textbf{88.06} \\ \hline
\end{tabular}

    % \vspace{-5mm}

\label{abla}
\end{table}

\section{Conclusion}
In this paper, we propose CalibraEval to mitigate the selection bias present in LLM-as-judges.
We reformulate the debiasing problem as an optimization problem and utilize the characteristics of unbiased evaluators as our optimization objectives. Moreover, we propose the non-parametric order-preserving algorithm (NOA) to determine the calibration function.
Experiments involving six LLMs across three representative datasets demonstrate that CalibraEval effectively reduces selection bias while enhancing accuracy. We argue that mitigating selection bias is essential for developing more reliable LLM evaluators. In the future, we plan to investigate additional biases in LLM-as-judges applications to create even more robust and trustworthy automated evaluations of large models.

% \clearpage
% \balance
\bibliographystyle{ACM-Reference-Format}
\bibliography{sample-base.bib}

\clearpage

\appendix

\section{SUPPLEMENTARY PROOF}
\label{proof}
To compute  $\frac{\partial L}{\partial d_{k}}$ for the given loss function:
\begin{equation}\label{eqn-20} 
[g(s_{0}^{i}) + g(s_2^i)-1]^2+[ g(s_{0}^{i}) - g(s_1^i)]^2-\lambda[ g(s_{0}^{i}) - g(s_2^i)]^2 
\end{equation}

Applying the chain rule, we have:

\begin{equation}\label{eqn-21} 
\frac{\partial L}{\partial d_{k}}=\frac{\partial L}{\partial g\left(s_{0}^{i}\right)} \frac{\partial g\left(s_{0}^{i}\right)}{\partial d_{k}}+\frac{\partial L}{\partial g\left(s_{1}^{i}\right)} \frac{\partial g\left(s_{1}^{i}\right)}{\partial d_{k}}+\frac{\partial L}{\partial g\left(s_{2}^{i}\right)} \frac{\partial g\left(s_{2}^{i}\right)}{\partial d_{k}}
\end{equation}

For $g(s_0^i)$:
\begin{equation}\label{eqn-22} 
\frac{\partial L}{\partial g\left(s_{0}^{i}\right)}=2\left[g\left(s_{0}^{i}\right)+g\left(s_{2}^{i}\right)-1\right]+2\left[g\left(s_{0}^{i}\right)-g\left(s_{1}^{i}\right)\right]
\end{equation}

For $g(s_1^i)$:
\begin{equation}\label{eqn-23} 
\frac{\partial L}{\partial g\left(s_{1}^{i}\right)}=-2\left[g\left(s_{0}^{i}\right)-g\left(s_{1}^{i}\right)\right]
\end{equation}

For $g(s_2^i)$:
\begin{equation}\label{eqn-24} 
\frac{\partial L}{\partial g\left(s_{2}^{i}\right)}=2\left[g\left(s_{0}^{i}\right)+g\left(s_{2}^{i}\right)-1\right]-2 \lambda\left[g\left(s_{0}^{i}\right)-g\left(s_{2}^{i}\right)\right]
\end{equation}

Then, We substitute the derivatives back into the equation \ref{eqn-21}:

\begin{align}\label{eqn-25} 
\frac{\partial L}{\partial d_{k}} & = \left(2\left[g\left(s_{0}^{i}\right)+g\left(s_{2}^{i}\right)-1\right]+2\left[g\left(s_{0}^{i}\right)-g\left(s_{1}^{i}\right)\right]\right) \frac{\partial g\left(s_{0}^{i}\right)}{\partial d_{k}} \nonumber \\
& \quad + \left(-2\left[g\left(s_{0}^{i}\right)-g\left(s_{1}^{i}\right)\right]\right) \frac{\partial g\left(s_{1}^{i}\right)}{\partial d_{k}}  \nonumber \\
& \quad + \left(2\left[g\left(s_{0}^{i}\right)+g\left(s_{2}^{i}\right)-1\right] - 2 \lambda\left[g\left(s_{0}^{i}\right) - g\left(s_{2}^{i}\right)\right]\right) \frac{\partial g\left(s_{2}^{i}\right)}{\partial d_{k}} 
\end{align}

Next, using the quotient rule, the derivative of $g(z_j)$ with respect to 
$d_k$ is:

\begin{equation}\label{eqn-26} 
\frac{\partial g\left(z_{j}\right)}{\partial d_{k}}=\frac{\partial}{\partial d_{k}}\left(\frac{\sum_{i=0}^{j} \exp \left(d_{i}\right)}{\sum_{i=0}^{M} \exp \left(d_{i}\right)}\right)
\end{equation}

% For j<k g(t_j)中只有分子中存在d_k,则

For $j<k$,  $d_k$ affects the denominator:
\begin{equation}\label{eqn-25} 
\frac{\partial g\left(z_{j}\right)}{\partial d_{k}}=\frac{-\sum_{i=0}^{j} \exp \left(d_{i}\right) \exp \left(d_{k}\right)}{\left(\sum_{i=0}^{M} \exp \left(d_{i}\right)\right)^{2}}
\end{equation}

For $j>=k$, $d_k$ affects both the numerator and the denominator:
\begin{equation}\label{eqn-25} 
\frac{\partial g\left(z_{j}\right)}{\partial d_{k}}=\frac{\exp\left(d_{k}\right) \left( \sum_{i=0}^{M} \exp\left(d_{i}\right) - \sum_{i=0}^{j} \exp \left(d_{i}\right) \right)}{\left(\sum_{i=0}^{M} \exp\left(d_{i}\right)\right)^{2}}
\end{equation}

% \begin{equation}\label{eqn-12} 
% g(t_k) = \frac{\sum_{i=0}^{k}exp(d_i) }{  {\textstyle \sum_{i=0}^{M}} exp(d_i)}
% \end{equation}

The final formula is as follows.
\begin{equation} \label{eq:partial_g}
\frac{\partial g\left(z_{j}\right)}{\partial d_{k}} =
\begin{cases}
-\frac{\sum_{i=0}^{j} \exp \left(d_{i}\right) \exp \left(d_{k}\right)}{\left(\sum_{i=0}^{M} \exp \left(d_{i}\right)\right)^{2}} & (j < k) \\
\frac{\exp\left(d_{k}\right) \left( \sum_{i=0}^{M} \exp\left(d_{i}\right) - \sum_{i=0}^{j} \exp \left(d_{i}\right) \right)}{\left(\sum_{i=0}^{M} \exp\left(d_{i}\right)\right)^{2}} & (j \ge k)
\end{cases}
\end{equation}

% 由于 影响的是分子中的和，偏导数为：
% ，因为这些项不在分子中，因此偏导数为
%% 推导过程
%% 数据%%更多的结果

\begin{algorithm}[t]
	\caption{Calibration process of CalibraEval}
	\label{algo}
	\KwIn{Language model, test samples $\mathcal{D}$, estimate set size $K$, threshold $\epsilon$}
	\KwOut{Debiased Prediction $\mathcal{Y}$}
        
        Sample $K$ estimation samples from the test samples $\mathcal{D}$

       \For{each sample $i \in \{1,...,K\}$}{
            Generate the probabilities after exchanging option IDs and positions.
    
            Obtain the set $s^i = \{s^i_0, s^i_1, s^i_2\}$ 
        }

        Combine all sets $s$ to form a global set $S = \bigcup_{i=1}^K S^i$

       Sort $S$ in ascending order and append $z_0=0, z_{3K+1}=1$. Obtain the sequence $Z = \{z_0,z_1,...,z_{3K},z_{3K+1}\}$
        
        Initialize the parameter $d_k$ as $z_k$ for each $k$

    \While{$\sum_{i=0}^{3K+1}\bigtriangleup d_i > \epsilon $}{
   
    % Move the leaf nodes in $Q$ to set $A$.
    \For{$i = 0$ to $3K+1$}{
    Calculate $\frac{\partial L}{\partial d_{i}}$ using Equation (\ref{eqn-14}) and Equation (\ref{eqn-15})
    
    Update the $d_i$ using Equation (\ref{eqn-13})
    }
    }

    Standardize $d_i$ to satisfy $\sum_{i=0}^{3K+1}d_i = 0$

    Obtain the discrete mapping function $g(\cdot)$ by using Equation (\ref{eqn-12})

    Obtain the continuous calibration function $g^*(\cdot)$ by solving Equation (\ref{eqn-16})

    \For{$ q \in \mathcal{D}$}{
        Debias the model prediction with $g^*(\cdot)$

        Add the predicted answer to $\mathcal{Y}$    
    }

    \Return Debiased Prediction $\mathcal{Y}$
\end{algorithm}

\section{Details of Dataset and Metrics}
\label{details}

\begin{table*}[ht]
\caption{Statistics of benchmark datasets.}
\begin{tabular}{c|c|ccc|ccc|lc}
\hline
\multirow{2}{*}{Datasets}    & \multirow{2}{*}{Total\_num} & \multicolumn{3}{c|}{Avergae\_Length}                                  & \multicolumn{3}{c|}{Label\_num}                                      & \multicolumn{1}{c}{\multirow{2}{*}{Type}} & \multirow{2}{*}{Type\_num} \\
                             &                             & prompt                & answer\_a             & answer\_b             & first                 & second                & tie                  & \multicolumn{1}{c}{}                      &                            \\ \hline
\multirow{4}{*}{RewardBench} & \multirow{4}{*}{2985}       & \multirow{4}{*}{1771} & \multirow{4}{*}{667}  & \multirow{4}{*}{658}  & \multirow{4}{*}{1490} & \multirow{4}{*}{1495} & \multirow{4}{*}{0}   & Chat                                      & 358                        \\
                             &                             &                       &                       &                       &                       &                       &                      & Chat\_Hard                                & 456                        \\
                             &                             &                       &                       &                       &                       &                       &                      & Safety                                    & 739                        \\
                             &                             &                       &                       &                       &                       &                       &                      & Reasoning                                 & 1432                       \\ \hline
\multirow{2}{*}{MTBench}     & \multirow{2}{*}{3355}       & \multirow{2}{*}{4039} & \multirow{2}{*}{1524} & \multirow{2}{*}{1512} & \multirow{2}{*}{1293} & \multirow{2}{*}{1282} & \multirow{2}{*}{780} & Turn1                                     & 1689                       \\
                             &                             &                       &                       &                       &                       &                       &                      & Turn2                                     & 1666                       \\ \hline
PreferenceBench              & 1998                        & 2485                  & 886                   & 893                   & 980                   & 1018                  & 0                    & -                                         & -                          \\ \hline
\end{tabular}

\label{data_sata}
\end{table*}

\subsection{Details of Datasets}
\label{data_detail}
Table \ref{data_sata} presents the statistical of benchmarks. The average answer length of different options in each dataset is nearly identical, and the distribution of label categories is balanced. This design minimizes the potential influence of other biases on the evaluation results. Overall, the datasets exhibit a well-balanced difficulty distribution and are thoughtfully constructed, ensuring a fair and robust evaluation process.

\subsection{Details of Metrics}
\label{metric_detail}

\subsubsection{Reference-Free Metrics}

Fleiss's Kappa is a statistical measure used to assess the reliability of agreement between multiple raters. It is calculated using the formula:

\begin{equation} \label{eq:partial_1}
K=\frac{P_{o}-P_{e}}{1-P_{e}}
\end{equation}
where $P_o$ is the observed agreement among raters. $P_e$ is the expected agreement by chance. The value of Kappa ranges from -1 to 1, where values closer to 1 indicate strong agreement among raters, values around 0 suggest no agreement beyond chance, and negative values indicate systematic disagreement.

Intraclass Correlation Coefficient (ICC) is a measure of reliability that assesses the consistency or agreement of measurements made by different raters or instruments. In this paper, we report two specific Intraclass Correlation Coefficient (ICC) metrics: ICC(2,k) and ICC(3,k). ICC(2,k) measures the consistency of ratings from multiple raters for the same set of subjects under a random effects model, while ICC(3,k) assesses the consistency of ratings from specific and fixed raters for the same subjects under a fixed effects model. Both are useful for measuring the reliability and consistency of ratings.

The ICC(2,k) is calculated using the formula:
\begin{equation} \label{eq:partial_2}
\operatorname{ICC}(2, k)=\frac{\sigma_{B}^{2}-\sigma_{W}^{2}}{\sigma_{B}^{2}+(k-1) \sigma_{W}^{2}}
\end{equation}

The ICC(3,k) is calculated using the formula:
\begin{equation} \label{eq:partial_3}
\operatorname{ICC}(3, k)=\frac{\sigma_{B}^{2}-\sigma_{W}^{2}}{\sigma_{B}^{2}+k \cdot \sigma_{W}^{2}}
\end{equation}
where $\sigma_{B}^{2}$ is the variance between the subjects.  $\sigma_{W}^{2}$ is the variance within the subjects. $k$ is the number of raters.

\subsubsection{Reference-based Metrics}

The standard deviation of recalls (RStd) quantifies the variability in recall scores across different evaluations. It is calculated using the formula:

\begin{equation} \label{eq:partial_4}
RStd=\sqrt{\frac{1}{N-1} \sum_{i=1}^{N}\left(R_{i}-\bar{R}\right)^{2}}
\end{equation}
where $R_{i}$ is the recall for the $i$-th evaluation. $\bar{R}$ is the mean recall across all evaluations. $N$ is the total number of evaluations.

Accuracy is a widely used evaluation metric that measures the overall correctness of a model's predictions:
\begin{equation} \label{eq:partial_5}
\text{Accuracy}=\frac{TP+TN}{TP+TN+FP+FN}
\end{equation}
where $TP$ represents the number of instances that are correctly predicted as positive, while $TN$ denotes the number of instances that are correctly predicted as negative. Conversely, $FP$ indicates the number of instances that are incorrectly predicted as positive, and $FN$ refers to the number of instances that are incorrectly predicted as negative.

\section{More Evaluation Result}
\label{more_results}
In Table \ref{refer-metric}, we present the complete results of the reference-based experimental metrics. On average, CalibraEval achieved better improvements in Rstd and accuracy. This indicates that CalibraEval effectively reduces selection bias and enables the model to realize its potential. By mitigating selection bias in the evaluation process, CalibraEval contributes to achieving more accurate and reliable results, paving the way for further advancements in model calibration and evaluation methodologies.

\section{THE DESIGN OF PROMPT}
\label{prompt}

Table \ref{more-prompt} presents all the prompts used in this paper. The default prompt, employed in the main experiments, serves as the foundational basis for assessing model performance. In the robustness experiments, four distinct prompts are utilized to evaluate variations in model responses.

\begin{table*}[t]
\vspace{-3mm}
\caption{
The complete results of reference-based metrics. We report the Standard Deviation of Recalls (RStd) and Accuracy (Acc.), with the best results highlighted in bold. $\downarrow$ indicates that lower values correspond to better performance.}
\vspace{-3mm}
\begin{tabular}{c|cc|cc|cc|cc}
\hline
\multirow{2}{*}{Model} & \multicolumn{2}{c|}{RewardBench} & \multicolumn{2}{c|}{MTBench}   & \multicolumn{2}{c|}{PreferenceBench} & \multicolumn{2}{c}{Average}    \\
                       & Rstd $\downarrow$           & Acc.(\%)       & Rstd$\downarrow$          & Acc.(\%)       & Rstd $\downarrow$            & Acc.(\%)          & Rstd $\downarrow$          & Acc.(\%)       \\ \hline
Llama-3-8B             & 15.01           & 65.79          & 16.42         & 67.08          & 3.36              & 83.43             & 11.60         & 72.10          \\
DI                     & 15.61           & 66.35          & 9.42          & 66.79          & 4.03              & 83.45             & 9.69          & 72.20          \\
CC                     & 14.62           & 64.52          & 8.70          & 69.09          & 9.47              & 82.78             & 10.93         & 72.13          \\
DC                     & 13.79           & 66.31          & 20.86         & 64.60          & 2.90              & 83.65             & 12.52         & 71.52          \\
Pride                  & 7.51            & 66.54          & 11.64         & 70.63          & 4.35              & 83.24             & 7.83          & 73.47          \\
CalibraEval            & \textbf{6.48}   & \textbf{68.12} & \textbf{5.22} & \textbf{70.63} & \textbf{2.42}     & \textbf{83.98}    & \textbf{5.04} & \textbf{74.24} \\ \hline
Llama-3.1-8B           & 17.93           & 64.96          & 14.73         & 67.58          & 6.65              & 77.54             & 13.10         & 70.03          \\
DI                     & 12.02  & 64.25          & 13.72         & 67.47          & 12.42             & 75.14             & 12.72         & 68.95          \\
CC                     & 8.43            & 65.39          & 6.75          & 65.09          & 6.31              & 77.91             & 7.16          & 69.49          \\
DC                     & 14.54           & 66.90          & 9.72          & 67.74          & \textbf{5.91}     & 77.98             & 10.06         & 70.87          \\
Pride                  & 13.94           & 65.90          & 12.63         & 67.56          & 9.47              & 77.92             & 12.01         & 70.46          \\
CalibraEval            & \textbf{6.88}   & \textbf{67.11} & \textbf{6.67} & \textbf{67.86} & 6.19              & \textbf{78.64}    & \textbf{6.58} & \textbf{71.20} \\ \hline
Qwen-14B               & 11.63           & 63.14          & 17.24         & 65.61          & 11.99             & 80.68             & 13.62         & 69.81          \\
DI                     & 9.76            & 61.88          & 19.09         & 62.18          & 15.77             & 76.14             & 14.87         & 66.73          \\
CC                     & 7.01            & 60.21          & 26.47         & 58.21          & 7.03              & 78.41             & 13.50         & 65.61          \\
DC                     & 3.02            & 62.47          & 8.23          & 68.48          & 10.07             & 79.96             & 7.11          & 70.30          \\
Pride                  & 4.18            & 64.09          & 16.31         & 65.29          & 7.36              & 83.55             & 9.28          & 70.98          \\
CalibraEval            & \textbf{2.72}   & \textbf{64.25} & \textbf{6.26} & \textbf{68.64} & \textbf{5.12}     & \textbf{83.88}    & \textbf{4.70} & \textbf{72.26} \\ \hline
Qwen-72B               & 4.01            & 87.20          & 5.76          & \textbf{81.32} & 2.54              & 90.12             & 4.10          & 86.21          \\
DI                     & 3.65            & 86.32          & 5.79          & 80.82          & 0.80              & 90.21             & 3.41          & 85.78          \\
CC                     & 7.72            & 85.74          & 4.72          & 81.05          & 5.41              & 89.30             & 5.95          & 85.36          \\
DC                     & 6.57            & 83.87          & 6.76          & 80.52          & 7.04              & 88.83             & 6.79          & 84.41          \\
Pride                  & 3.82            & 87.25          & 5.24          & 81.23          & 2.27              & 90.26             & 3.78          & 86.25          \\
CalibraEval            & \textbf{0.94}   & \textbf{88.06} & \textbf{4.99} & 81.25          & \textbf{0.69}     & \textbf{90.71}    & \textbf{2.21} & \textbf{86.67} \\ \hline
ChatGPT                & 16.79           & 65.27          & 7.66          & 72.67          & 3.04              & 85.61             & 9.16          & 74.70          \\
DI                     & 7.22            & 65.24          & 7.01          & 69.84          & 9.82              & 83.94             & 8.02          & 73.01          \\
CC                     & 7.93            & 64.89          & 16.31         & 70.49          & 3.13              & 84.83             & 9.12          & 73.40          \\
DC                     & 11.40           & 66.89          & 20.23         & 68.67          & 5.81              & 82.68             & 12.48         & 72.75          \\
Pride                  & 8.54            & 66.36          & 6.01          & 72.86          & 3.51              & 85.68             & 6.02          & 74.97          \\
CalibraEval            & \textbf{5.51}   & \textbf{67.13} & \textbf{5.20} & \textbf{72.98} & \textbf{2.82}     & \textbf{85.98}    & \textbf{4.51} & \textbf{75.36} \\ \hline
GPT4o                  & 1.95            & 89.34          & 3.23          & 82.27          & 5.29              & 90.44             & 3.49          & 87.35          \\
DI                     & 3.91            & 88.64          & 2.24          & 82.26          & 4.49              & 90.46             & 3.55          & 87.12          \\
CC                     & \textbf{0.54}   & 89.11          & 4.84          & 81.23          & 5.79              & 90.02             & 3.72          & 86.79          \\
DC                     & 1.84            & 87.84          & 3.57          & 81.93          & 5.99              & 88.22             & 3.80          & 86.00          \\
Pride                  & 1.68            & 89.38          & 3.84          & 82.10          & 5.24              & 90.47             & 3.60          & 87.32          \\
CalibraEval            & 1.42            & \textbf{89.54} & \textbf{2.89} & \textbf{82.29} & \textbf{4.29}     & \textbf{90.49}    & \textbf{2.87} & \textbf{87.44} \\ \hline
\end{tabular}
\label{refer-metric}
\end{table*}

\begin{table*}[t]
\label{more-prompt}
\caption{Different promtp templates used in this paper}
\vspace{-3mm}
\begin{tabular}[c]{p{0.95\textwidth}}
\hline
\textbf{Default Prompt.}                                                                                                                                                                                                   \\ \hline
  \begin{tabular}[c]{@{\ }p{0.95\textwidth}@{\ }}Given a question and two answers. Determine which one better answers the question. You only need to output A or B directly to indicate which answer is better.\end{tabular}                            \\ \hline

\textbf{Prompt Variant One.}                                                                                                                                                                                                            \\ \hline
  \begin{tabular}[c]{@{\ }p{0.95\textwidth}@{\ }}Please evaluate the quality of the responses to the question displayed below.  Don't provide your explanation, only output your final verdict by strictly following this format: A if assistant A is better, B if assistant B is better.\end{tabular} \\ \hline

\textbf{Prompt Variant Two.}                                                                                                                                                                                                            \\ \hline
  \begin{tabular}[c]{@{\ }p{0.95\textwidth}@{\ }}You are an advanced evaluator, and your task is to assess which response addresses the inquiry more effectively. Output A if response A is better, or B if response B is better.\end{tabular} \\ \hline

\textbf{Prompt Variant Three.}   
\\ \hline
  \begin{tabular}[c]{@{\ }p{0.95\textwidth}@{\ }}Below is a query along with two different responses generated by AI assistants. Your task is to determine which response provides a more accurate and helpful answer to the question posed. Don't provide your explanation. Simply output A if response A is more effective, or B if response B is more effective.\end{tabular} \\ \hline

\end{tabular}
\label{more-prompt}
\end{table*}

\end{document}